\documentclass[10pt, conference, compsoc]{IEEEtran}

\usepackage{amssymb}
\usepackage{latexsym}
\usepackage{url}
\usepackage{graphicx}
\usepackage{float}
\usepackage[labelfont=bf]{caption}
\usepackage{color}
\usepackage{xcolor}
\usepackage{subcaption}
\usepackage{hyperref}
\usepackage{siunitx}
\usepackage{booktabs}
\usepackage{rotating}

\usepackage{framed,multirow}
\usepackage{amsmath}
\usepackage{array}
\usepackage{stackengine}
\usepackage{relsize}
\usepackage[inline]{enumitem}
\usepackage[ruled,vlined]{algorithm2e}
\usepackage{makecell}
\usepackage{pifont}
\usepackage{amsthm}

\usepackage{stfloats}

\newcommand{\ie}{\textit{i}.\textit{e}., }
\newcommand{\eg}{\textit{e}.\textit{g}., }

\newcommand{\cmark}{\ding{51}}%
\newcommand{\xmark}{\ding{55}}

\makeatletter
\newcommand\notsotiny{\@setfontsize\notsotiny{6}{7}}
\makeatother

\newcommand{\diff}[1]{{\textcolor{red}{\notsotiny($#1$)}}}

\usepackage[capitalize]{cleveref}
\crefname{section}{Sec.}{Secs.}
\Crefname{section}{Section}{Sections}
\Crefname{table}{Table}{Tables}
\crefname{table}{Tab.}{Tabs.}

\begin{document}
\title{Generative appearance replay for continual unsupervised domain adaptation}

\author{
\IEEEauthorblockN{Boqi Chen$^{1,2}$*,
Kevin Thandiackal$^{1,2}$*,
Pushpak Pati$^{1}$ and
Orcun Goksel$^{2,3}$}\\
\IEEEauthorblockA{$^{1}$IBM Research Europe, Zurich, Switzerland}
\IEEEauthorblockA{$^{2}$Computer-assisted Applications in Medicine, ETH Zurich, Zurich, Switzerland}
\IEEEauthorblockA{$^{3}$Department of Information Technology, Uppsala University, Uppsala, Sweden}
}

\maketitle

\makeatletter%
\long\def\@makefntext#1{%
  \noindent \hb@xt@ 1.8em{\hss\@makefnmark}#1}
\makeatother

\begingroup\renewcommand\thefootnote{*}
\footnotetext{The authors contributed equally to this work.}
\endgroup

\begin{abstract}
Deep learning models can achieve high accuracy when trained on large amounts of labeled data.
However, real-world scenarios often involve several challenges:
Training data may become available in installments, may originate from multiple different domains, and may not contain labels for training.
Certain settings, for instance medical applications, often involve further restrictions that prohibit retention of previously seen data due to privacy regulations.
In this work, to address such challenges, we study unsupervised segmentation in continual learning scenarios that involve domain shift.
To that end, we introduce GarDA (Generative Appearance Replay for continual Domain Adaptation), a generative-replay based approach that can adapt a segmentation model sequentially to new domains with unlabeled data.
In contrast to single-step unsupervised domain adaptation (UDA), continual adaptation to a sequence of domains enables leveraging and consolidation of information from multiple domains.
Unlike previous approaches in incremental UDA, our method does not require access to previously seen data, making it applicable in many practical scenarios.
We evaluate GarDA on two datasets with different organs and modalities, where it substantially outperforms existing techniques.
\end{abstract}


\IEEEpeerreviewmaketitle

\section{Introduction}
\label{sec:intro}

Deep Neural Networks (DNNs) have recently achieved remarkable performance on various computer vision tasks with natural images, such as classification~\cite{he2016resnet,dosovitskiy2020vit,liu2022convnext} and semantic segmentation~\cite{long2015fcn,chen2017deeplab}.
However, there exist several challenges that hinder DNNs from achieving similar success in other domains, \eg healthcare.
First, to achieve high-performance, DNNs require large amounts of labeled training images, which are challenging to obtain for medical applications, since annotations can only be provided by medical experts.
Annotating medical images is therefore more costly compared to annotating natural images.
This is particularly critical for applications requiring dense annotations, such as semantic segmentation.
Thus, there is a strong need for \emph{unsupervised} DNN approaches in healthcare.
Second, medical datasets usually contain a relatively small number of images~\cite{porwal2018idrid,batista2020rimone,campello2021mnms,brancati2022bracs} compared to large-scale natural image datasets, such as ImageNet~\cite{russakovsky2015imagenet}.
Models trained on such small datasets often do not generalize well to unseen domains~\cite{vanderlaak2021dlhistosurvey}. 
Unsupervised Domain Adaptation (UDA)~\cite{kouw2019udareview,wilson2020udasurvey} aims to address both aforementioned challenges, by first training a model on a \emph{labeled source} domain $\mathcal{D}_0$ and afterwards adapting it to an \emph{unlabeled target} domain $\mathcal{D}_1$, to alleviate annotation costs.

Real-world settings are often more complex than the above scenario, and require long-term applicable solutions rather than adapting to a single target domain alone.
It is, for example, typical that multiple datasets from different domains $\mathcal{D}_1, ..., \mathcal{D}_T$ become available at different time points.
In such a scenario, it is beneficial to keep a single model that is sequentially adapted to these different domains, thereby consolidating useful knowledge from more than one domain.
Since patient data is bound to strict privacy regulations, data sharing and indefinite data storage are often not possible.
Therefore such medical scenarios are further complicated by the inability to keep and access data from previously seen domains.
In the literature, this constrained learning problem is known as continual learning (CL).
In this work, we study a \emph{continual} UDA setting, where a model is adapted to a sequence of target domains, the data of which is available only while training on that respective domain for the first time.

\begin{figure}
    \centering
    \includegraphics[width=\linewidth]{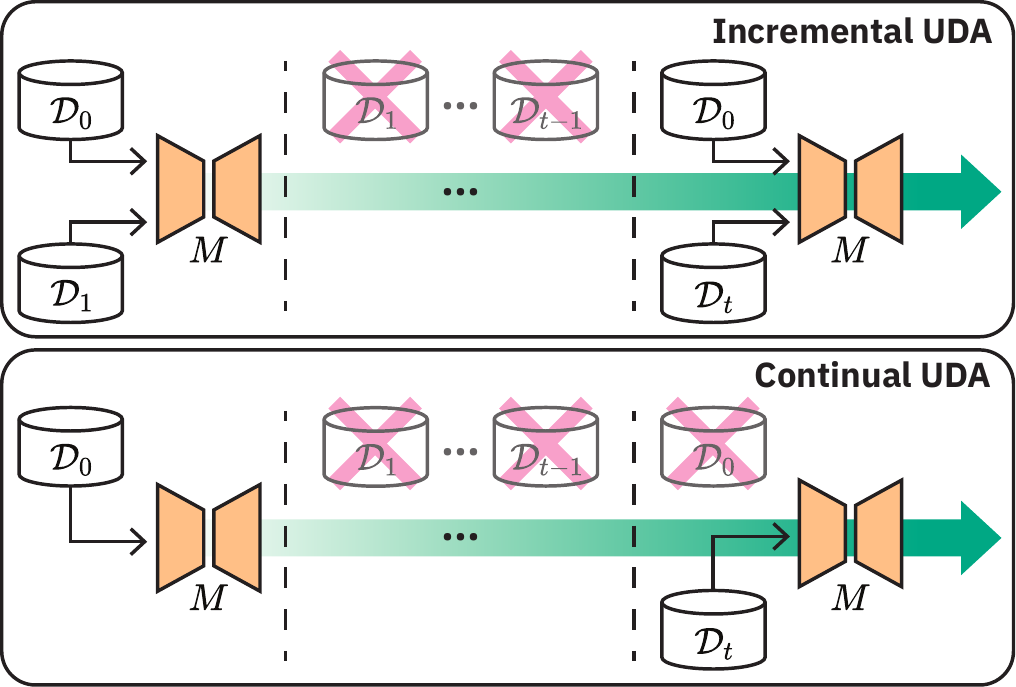}
    \caption{Overview of incremental (top) and continual (bottom) UDA workflows. In the latter, the segmentation model $M$ sequentially adapts to new domains without storing data from \emph{any} previously seen domain, including $\mathcal{D}_0$.}
    \label{fig:teaser}
\end{figure}

The main challenges in continual UDA are twofold.
First, the model should not suffer from catastrophic forgetting~\cite{mccloskey1989forgetting,parisi2019clreview}, \ie it should maintain its performance on the source domain as well as all previously seen target domains.
Second, without access to the source or previous target domains, the model should adapt accurately to new target domains.
Although strict continual UDA has been largely unexplored in the literature, some recent techniques have studied a scenario with looser constraints, which we herein term as \emph{incremental} UDA.
In contrast to continual UDA, in incremental UDA the data from the source domain is stored and reused during the introduction of each subsequent target domain, as illustrated in~\cref{fig:teaser}.

As an incremental UDA method, \cite{saporta2022multi} proposed Multi-Head Distillation (MuHDi), which performs  knowledge distillation using either the source images (which are kept at all times) or the images from the current target domain.
This is sub-optimal since knowledge from previously seen target domains cannot be preserved effectively.
Another method, more closely related to our solution, is ACE (Adapting to Changing Environments)~\cite{wu2019ace}, which builds on the idea of transferring the style of unlabeled target images to known labeled source images (which then again have to be kept indefinitely).
To address forgetting, ACE generates images with the same content, \ie segmentation ground truth, as a given source image, but the style of a target image.
In addition, ACE employs a memory unit to replay images in previously seen target styles to the segmentation model.
However, as for MuHDi, ACE also violates a strict continual UDA constraint, since it relies on future access to the source data.
Furthermore, ACE can only replay the previous target styles that have been stored in the memory unit, 
hence lacking the ability to sample those for replay from the \emph{entire} distribution of previously seen target styles.
This in turn affects the knowledge preservation from these intermediate targets.

In this work, we address the above-mentioned shortcomings by proposing GarDA (Generative Appearance Replay for continual Domain Adaptation).
We employ generative replay~\cite{Shin2017ContinualReplay,Kemker2017FearNet:Learning,CongGANForgetting,thandiackal2021genifer} to substitute images from any previously-seen but currently-inaccessible domain with synthetic images that are generated via a generative adversarial network (GAN)~\cite{Goodfellow2014gans}.
This permits operation under a strict continual UDA setting, \ie even when the source and the intermediate target domains become inaccessible after training on each.
Importantly, for replay we can arbitrarily sample images of any seen domain via our generator, and are not limited by less expressive and fixed memory buffers, as in ACE.
Furthermore, thanks to our powerful GAN-based approach for transferring the appearance of target domain images onto source domain images, we can generate more representative samples for replay, achieving more effective knowledge preservation.

In summary, our main contributions are three-fold: 
(1)~We propose GarDA as the first segmentation technique that operates under a strict continual UDA setting, \ie it mitigates forgetting of the source and intermediate target domains without storing \emph{any} previously seen images.
(2)~We introduce a novel combination of appearance transfer and generative replay using a stochastic generator, which enables diverse sampling from the entire distribution of all domains, and is thus less prone to overfitting compared to methods employing separate memory buffers.
(3)~We achieve the new state-of-the-art performance in continual UDA, demonstrated herein through comprehensive benchmarking with two target domains. 
To demonstrate the generalizability of GarDA, we conduct experiments for two entirely different tasks: optic disc segmentation in color fundus photography (CFP) images and cardiac segmentation in magnetic resonance (MR) images.

\section{Related work}
\label{sec:related}
Our work lies at the intersection of UDA and continual learning, for which we provide below a brief overview with relevant research.

\subsection{Continual learning}
Continual learning aims to simulate real-world data availability constraints in sequential learning, \ie a model should learn from a sequence of datasets without retaining any previously seen data and without forgetting previously acquired knowledge~\cite{Farquhar2018TowardsLearning,parisi2019clreview,vandeVen2022ThreeLearning,DeLange2021ATasks}.
The research in continual learning can be categorized into four major dimensions focusing on regularization~\cite{Kirkpatrick2017OvercomingNetworks,Zenke2017ContinualIntelligence,Nguyen2018VariationalLearning}, parameter isolation~\cite{Mallya2018Piggyback,Serra2018OvercomingTask,Wortsman2020SupermasksSuperposition}, prototypical representations~\cite{yu2020sdc,Zhu2021Pass,toldo2022Fusion}, and generative replay~\cite{Shin2017ContinualReplay,Kemker2017FearNet:Learning,WuMemoryForgetting,OstapenkoLearningLearning,Liu2020GenerativeLearning,vandeVen2020Brain-inspiredNetworks,CongGANForgetting,thandiackal2021genifer}.
Methods built on prototypical representations as well as generative replay have been the most successful.
The former aims to store class-representative prototypes in feature space to compensate any estimated drift when learning new tasks.
In the latter, a generative model learns to synthesize images~\cite{Shin2017ContinualReplay,WuMemoryForgetting,OstapenkoLearningLearning,CongGANForgetting,thandiackal2021genifer} or features~\cite{Kemker2017FearNet:Learning,Liu2020GenerativeLearning,vandeVen2020Brain-inspiredNetworks} of previously seen data, such that they can be replayed to the classification/segmentation model when the original data is not available anymore.
While these methods have achieved state-of-the-art performance on class- and task-incremental benchmarks, they have not been employed in the context of continual UDA, to the best of our knowledge.

\subsection{UDA}
Over the last years, a myriad of methods have been proposed to tackle UDA for both classification and segmentation~\cite{kouw2019udareview,wilson2020udasurvey}, with their common paradigm often being the domain alignment.
This aims to bring the source and target domain representations closer, in terms of either the input images, certain intermediate features, or the model outputs.
If this is done effectively, the classification/segmentation models trained on the source domain are expected to generalize to the target domain.
\cite{ganin2016dann} were the first to propose a Domain-Adversarial Neural Network (DANN), where the main model's feature extractor uses the negative gradient of a domain classifier to obtain domain-invariant features.
Similarly, other approaches have adopted GAN-inspired adversarial training for feature alignment~\cite{tzeng2017adda,long2018cdan,xu2019afn}, which has been effective also for the segmentation of medical images such as MR and computer tomography (CT) images~\cite{kamnitsas2017unsupervised,dou2019pnpadanet}.
Other methods such as AdaptSegNet~\cite{tsai2018adaptsegnet} or AdvEnt~\cite{vu2019advent} also employ adversarial losses, but these aim at aligning the model output spaces of source and target domains.
Alternatively, CycleGAN~\cite{zhu2017cyclegan} aims to align domains in the image space.
Cycle-Consistent Adversarial Domain Adaptation (CyCADA)~\cite{hoffman2018cycada} and Synergistic Image and Feature Alignment (SIFA)~\cite{chen2020sifa} propose to unify feature-level and image-level domain alignment for UDA.

All the aforementioned UDA methods were designed to adapt models from a single source domain to a single target domain, and they do not operate in continual (or incremental) learning settings.
In this work, our goal is to tackle continual learning settings, for reasons motivated earlier including the limitations of access to medical data.
Continual UDA, with a sequence of multiple target domains, introduces additional challenges to single-domain UDA.

\subsection{Sequential UDA}
In the literature, UDA for a sequence of multiple target domains has been studied under different settings.
We herein group them collectively under the umbrella of \emph{sequential UDA}, and propose the taxonomy presented in~\Cref{tab:da_settings} to clarify the differences among the individual settings.
\begin{table}
    \setlength{\tabcolsep}{2pt} 
    \footnotesize
    \centering
    \caption{Taxonomy of different domain adaptation settings.}
    \begin{tabular}{lccc}
        \toprule
        Setting & \thead{Source \\unavailable \\after training} & \thead{Multiple \\target \\domains} & \thead{Unlabeled \\target \\domain(s)} \\
        \midrule
        Unsup.\ Domain Adaptation (UDA) & \xmark & \xmark & \cmark \\
        \midrule
        Domain-Incremental Learning (DIL) & \cmark & \cmark & \xmark \\
        Source-free UDA / UMA & \cmark & \xmark & \cmark \\
        Incremental UDA & \xmark & \cmark & \cmark \\
        Continual UDA & \cmark & \cmark & \cmark \\
        \bottomrule
    \end{tabular}
    \label{tab:da_settings}
\end{table}
We denote the problem of learning multiple \emph{labeled} target domains without retaining previously seen data as domain-incremental learning  (DIL)~\cite{karani2018lifelong,lenga2020clxray,gonzalez2020wrongcl,srivastava2021cdilxray,kalb2021cildil,li2022dilstylereplay,garg2022multi,ranem2022continual}.
Another sub-category called source-free UDA -- sometimes referred to as unsupervised model adaptation (UMA) -- has recently gained increasing attention for both classification~\cite{kundu2020universal,li2020uma,yang2021generalizedsfda,xia2021adaptivesfda} and segmentation tasks~\cite{liu2021sfdaseg,huang2021umacontrastive,kundu2021generalize,stan2021uma,chen2021sfdafundus}.
These methods perform UDA from a source to a single target domain without using the actual source data.
Note that in the literature, source-free and continual UDA have sometimes been used interchangeably.
Instead, we propose to separate these terms as shown in~\Cref{tab:da_settings}, since source-free UDA aims for adaptation to one target domain only, therefore not having to consider resilience during further target adaptations in the future.
Naming this setting as continual UDA would be indeed inconsistent with the remaining continual learning literature~\cite{parisi2019clreview,DeLange2021ATasks} that deals with scenarios involving multiple classes/tasks in a sequence.

In contrast to the above, incremental UDA aims at adapting models to multiple target domains in an unsupervised manner.
For instance, MuHDi~\cite{saporta2022multi} employs domain adversarial training for adaptation, together with multiple distillation losses to mitigate forgetting.
As mentioned earlier, knowledge distillation is sub-optimal when enforced using the source images and/or the current target images, as these do not directly prevent forgetting of the previously seen target domains.
Wu et al.~\cite{wu2019ace} proposed ACE, which uses a deterministic encoder-decoder architecture to create images in a target domain style for replay.
To tackle forgetting, ACE employs a replay memory buffer containing style information of previously seen target domains.
Although this can mitigate forgetting to some extent, the memory buffer can only hold the style information of few samples, thus not representative of the complete distribution of all previous target domains/images.
In addition, both MuHDi and ACE require access to source data throughout the adaptation sequence, making them incremental UDA in our taxonomy.

In this work, we study continual UDA, a more challenging scenario compared to those aforementioned:
The source data is available only once at the beginning of the sequence for training on the source domain.
Afterwards, the model adapts to a sequence of unlabeled target domains without storing data from any previous domain.

\section{Method}
\label{sec:method}

In this section, we present our proposed framework for continual UDA.
\begin{figure*}
    \centering
    \includegraphics[width=0.9\linewidth]{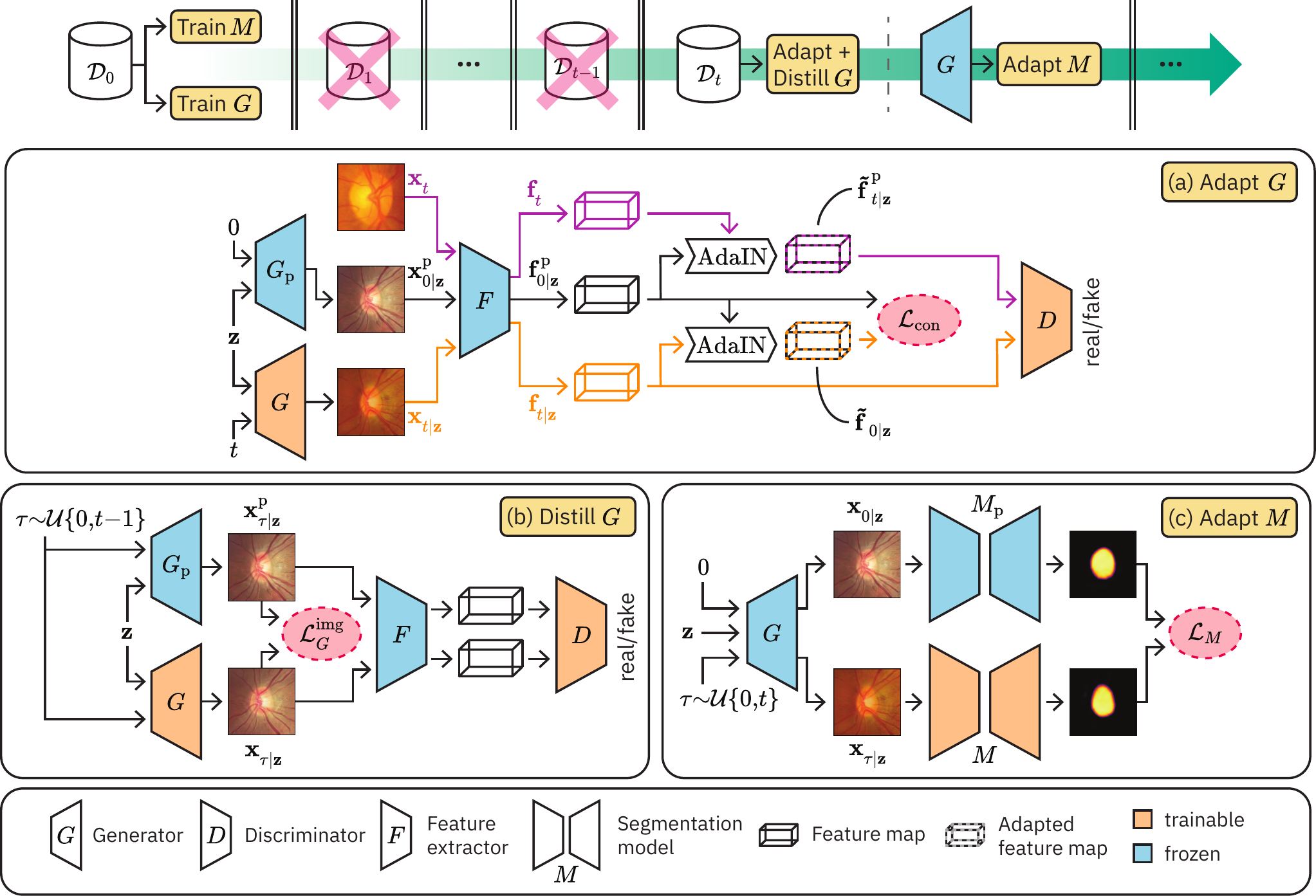}
    \caption{Overview of our method including (a) GAN adaptation (b) GAN distillation and (c) adaptation of the segmentation model.}
    \label{fig:overview}
\end{figure*}
A segmentation model $M$ is first trained on images $\mathbf{x}_0\in\mathcal{X}_0\subseteq\mathbb{R}^{H \times W \times C}$ with segmentation label maps $\mathbf{y}_0\in\mathcal{Y}_0\subseteq\{1, ..., L\}^{H \times W}$ from the source domain $\mathcal{D}_0=\{(\mathbf{x}_0, \mathbf{y}_0)^{(i)}\}_{i=1}^{N_0}$, where $N_0$, $L$, $C$, $H$, and $W$ denote the number of images, labels, image channels, image height, and image width, respectively.
Following training on the source, $\mathcal{D}_0$ is afterwards not accessible anymore while adapting $M$ to a sequence of $T\geq 1$ target domains $\mathcal{D}_{1:T}=\{ \mathcal{X}_1, ..., \mathcal{X}_T \}$ which are lacking labels.
In other words, while training on the $t$-th target domain, we only have access to the unlabeled images from the current domain, \ie $\mathbf{x}_t\in\mathcal{D}_t$, but we cannot access samples from the source or previously encountered target domains, \ie $\mathcal{D}_{0:t-1}$.
We substitute the missing $\mathcal{D}_{0:t-1}$ with synthetic data produced by a GAN and replay this data to $M$ to prevent forgetting.
As new target domains are added, the GAN (consisting of a generator $G$ and a discriminator $D$) and the segmentation model $M$ are trained alternately, as illustrated in~\Cref{fig:overview}.

First, $M$ and the GAN are trained independently on the source domain~(\Cref{sec:source_training}).
Next, $G$ is trained to synthesize images with the appearance characteristics of a new target domain~(\Cref{fig:overview}(a)) while remembering how to generate images of previous domains~(\Cref{fig:overview}(b)).
This training phase is described in~\Cref{sec:gan_da} and \Cref{sec:gan_dist}, respectively.
From such a generator $G$, at any point we can create pairs of synthetic samples with the same \emph{content} but with separate appearances: one of the source domain and one of a target domain.
We can then use these pairs to \emph{adapt} the segmentation model, by distilling the outputs of the previously trained model $M_\mathrm{p}$ into the current model $M$, as seen in \Cref{fig:overview}(c).
In other words, the new $M$ shall predict the same segmentation map for a target-like image as the map predicted by $M_\mathrm{p}$ for the corresponding source-like image.
This step is described in~\Cref{sec:seg_training}.
The above process is repeated whenever data from a new target domain becomes available.

\subsection{Source training}
\label{sec:source_training}

Initially, we train a segmentation model $M$ and a GAN ($G$, $D$) separately on the source domain data. 
For an input image $\mathbf{x}$, $M$ predicts a segmentation map.
Specifically, we denote the output logit of $M$ for label $l$ at pixel position $(h, w)$ as $M^{(h, w, l)}(\mathbf{x})$.
The corresponding probability of belonging to label $l$ is computed via a softmax operation, \ie
\begin{equation}
    m^{(h, w, l)}(\mathbf{x}) = \frac{e^{M^{(h, w, l)}(\mathbf{x})}}{\sum_{j=1}^{L} e^{M^{(h, w, j)}(\mathbf{x})}} \;.
    \label{eq:softmax}
\end{equation}
Given a source image and segmentation label map $(\mathbf{x}, \mathbf{y})\in\mathcal{D}_0$, $M$ is trained by minimizing the cross-entropy loss
\begin{equation}
    \mathcal{L}_M\!= - \frac{1}{H W} \!\sum_{h=1}^{H} \sum_{w=1}^{W} \sum_{l=1}^{L} 1_{\{\mathbf{y}^{(h, w)} = l\}} \!\log \!\left( \!m^{(h, w, l)}(\mathbf{x}) \!\right)
    \label{eq:l_m_source}
\end{equation}
where $1_{\{\mathbf{y}^{(h, w)} = l\}}$ denotes an indicator function that is 1 if $\mathbf{y}^{(h, w)}$$=$$l$ and 0 otherwise.

For the generative replay model, we employ a conditional GAN consisting of a generator $G$ and a projection discriminator $D$~\cite{Miyato2018CGAN}.
Given a randomly sampled noise vector $\mathbf{z}$ (that aims to condition the image content) and a domain label $\tau$ (for the corresponding appearance), $G:(\tau, \mathbf{z})\mapsto\mathbf{x}_{\tau|\mathbf{z}}$ learns to generate synthetic samples $\mathbf{x}_{\tau|\mathbf{z}}$ of domain $\mathcal{D}_\tau$.
$G$ is trained by optimizing a non-saturating logistic loss~\cite{Goodfellow2014gans}
\begin{equation}
    \mathcal{L}_G = \mathop{\mathbb{E}}_{\mathbf{z} \sim p_z} \Big[ a\Big(-D\big(0, G(0, \mathbf{z})\big) \Big) \Big]\;,
    \label{eq:l_g_source}
\end{equation}
where $a(\cdot)$ is the softplus operation.
During training, $D$ counteracts $G$ by trying to distinguish synthetic from real images, by minimizing
\begin{equation}
    \begin{aligned}
    \mathcal{L}_{D} =& \mathop{\mathbb{E}}_{\mathbf{z} \sim p_z} \Big[ a\Big( D \big( 0, G(0, \mathbf{z}) \big) \Big) \Big]\\
    &+ \mathop{\mathbb{E}}_{\mathbf{x} \sim \mathcal{D}_0} \Big[ a\Big(-D \big(0,\mathbf{x} \big) \Big) \Big] + \lambda_{R_1} R_1 \;,
    \label{eq:l_d_source}
    \end{aligned}
\end{equation}
where $R_1$ is the gradient penalty term $\mathop{\mathbb{E}}_{\mathbf{x} \sim \mathcal{D}_0} \left\lVert \nabla_{\mathbf{x}} D(0, \mathbf{x}) \right\rVert^2_2$ \cite{Mescheder2018R1}, which is only computed for real images and weighted with $\lambda_{R_1}$.

\subsection{GAN adaptation}
\label{sec:gan_da}

The main challenge following the source training is the unavailability of labeled target domain images, which prevents the trivial solution of fine-tuning $M$ on the target domain.
However, the challenge can be addressed, \eg if a mechanism could create (pseudo-)labeled target domain images.
To this end, we propose to adapt our GAN to the target domain such that synthetic images can be sampled with \emph{content} similar to the source and \emph{appearance} similar to any target.

As presented in~\Cref{fig:overview}(a), for a new target domain $\mathcal{D}_t$, we
freeze the previously trained generator $G_{0:t-1}$ and instantiate a trainable generator $G_t$. For simplicity, we denote $G_t$ being trained as $G$, and frozen $G_{0:t-1}$ for previous domains as $G_\mathrm{p}$. 
We initialize $G$ with the weights of $G_\mathrm{p}$.

We employ a discriminator $D$ that operates on \emph{feature representations} extracted using a frozen pretrained feature extractor $F$, \ie instead of distinguishing between real and fake \emph{images}, our $D$ distinguishes their feature representations.
Accordingly, we first randomly sample a source-like image $\mathbf{x}^\mathrm{p}_{0|\mathbf{z}}$$=$$G_\mathrm{p}(0, \mathbf{z})$ from $G_\mathrm{p}$ and a real image $\mathbf{x}_t$ from the current dataset $\mathcal{D}_t$.
These images are then passed through $F$, yielding feature maps $\mathbf{f}^\mathrm{p}_{0|\mathbf{z}}$$=$$F(\mathbf{x}^\mathrm{p}_{0|\mathbf{z}})$, and $\mathbf{f}^{\vphantom{\mathrm{p}}}_{t\vphantom{|}}$$=$$F(\mathbf{x}^{\vphantom{\mathrm{p}}}_{t\vphantom{|}})$.
We now aim to create a target-like image $\mathbf{x}^{\vphantom{\mathrm{p}}}_{t|\mathbf{z}}$$=$$G(t, \mathbf{z})$ with features $\mathbf{f}^{\vphantom{\mathrm{p}}}_{t|\mathbf{z}}$$=$$F(\mathbf{x}^{\vphantom{\mathrm{p}}}_{t|\mathbf{z}})$ that incorporates the content of $\mathbf{x}^\mathrm{p}_{0|\mathbf{z}}$ and the appearance of $\mathbf{x}^{\vphantom{\mathrm{p}}}_{t\vphantom{|}}$.
In order to train $G$, we first need to define an objective, \ie what the synthesized features $\mathbf{f}^{\vphantom{\mathrm{p}}}_{t|\mathbf{z}}$ should look like.

Inspired by~\cite{huang2017adain,wu2019ace}, we leverage the idea behind adaptive instance normalization (AdaIN) that the instance-level statistics (\eg mean and variance) of a feature map represent the appearance of the corresponding image.
Using AdaIN, we can re-normalize the instance-level statistics of $\mathbf{f}^\mathrm{p}_{0|\mathbf{z}}$ to those of $\mathbf{f}^{\vphantom{\mathrm{p}}}_{t\vphantom{|}}$ and obtain a new feature map $\mathbf{\tilde{f}}^\mathrm{p}_{t|\mathbf{z}}$ that encodes the content of $\mathbf{x}^\mathrm{p}_{0|\mathbf{z}}$ and the appearance of $\mathbf{x}^{\vphantom{\mathrm{p}}}_{t\vphantom{|}}$.
The said re-normalization is computed as
\begin{equation}
    \mathbf{\tilde{f}}^\mathrm{p}_{t|\mathbf{z}}\!=\!\text{AdaIN}\!\left(\!\mathbf{f}^\mathrm{p}_{0|\mathbf{z}},\mathbf{f}^{\vphantom{\mathrm{p}}}_{t\vphantom{|}}\!\right)\!=\!\sigma(\mathbf{f}^{\vphantom{\mathrm{p}}}_{t\vphantom{|}})\!\!\left(\!\frac{\mathbf{f}^\mathrm{p}_{0|\mathbf{z}}-\mu\left(\!\mathbf{f}^\mathrm{p}_{0|\mathbf{z}}\!\right)}{\sigma\left(\!\mathbf{f}^\mathrm{p}_{0|\mathbf{z}}\!\right)}\!\right)\!+\mu(\mathbf{f}^{\vphantom{\mathrm{p}}}_{t\vphantom{|}}),
    \label{eq:adain}
\end{equation}
where $\mu(\cdot)$ and $\sigma(\cdot)$ denote channel-wise mean and standard deviation, respectively.
During GAN training, discriminator $D$ is trained to identify $\mathbf{\tilde{f}}^\mathrm{p}_{t|\mathbf{z}}$ as real and $\mathbf{f}^{\vphantom{\mathrm{p}}}_{t|\mathbf{z}}$ as fake.
To this end, $D$ optimizes the following objective:
\begin{equation}
    \begin{aligned}
    \mathcal{L}^{\mathrm{uda}}_{D} =& \mathop{\mathbb{E}}_{\mathbf{z} \sim p_z} \Big[ a \Big( D\Big(t, \mathbf{f}^{\vphantom{\mathrm{p}}}_{t|\mathbf{z}}\Big) \Big) \Big]\\
    &+ \mathop{\mathbb{E}}_{\substack{\mathbf{z} \sim p_z \\ \mathbf{x}_t \sim \mathcal{D}_t}} \Big[ a \left(-D\left(t, \mathbf{\tilde{f}}^\mathrm{p}_{t|\mathbf{z}}\right) \right) \Big] + \lambda_{R_1} R_1 \;.
    \label{eq:l_uda_d}
    \end{aligned}
\end{equation}
Meanwhile, $G$ tries to confuse $D$ while adapting to the new domain by minimizing
\begin{equation}
  \mathcal{L}^{\mathrm{uda}}_{G} = \mathop{\mathbb{E}}_{\mathbf{z}\in p_z} \Big[ a \Big( -D\left(t, \mathbf{f}^{\vphantom{\mathrm{p}}}_{t|\mathbf{z}}\right) \Big) \Big] + \lambda_\mathrm{con} \mathcal{L}_\mathrm{con}\;, \label{eq:l_uda_g}
\end{equation}
where the first term encourages $G$ to generate images with realistic target appearance and the second term is a content loss.
The purpose of the content loss is for $G$ to generate two images with different appearance characteristics, but the same content, when given the same $\mathbf{z}$ and two different domain labels as input.
That means that in the feature space, the only difference between $\mathbf{f}^{\vphantom{\mathrm{p}}}_{t|\mathbf{z}}$ and $\mathbf{f}^\mathrm{p}_{0|\mathbf{z}}$ should be the appearance information, which we estimate through channel-wise mean and variance.
Thus, we define a content loss as the difference between the synthesized target features after re-normalization $\mathbf{\tilde{f}}^{\vphantom{\mathrm{p}}}_{0|\mathbf{z}}$$=$$\text{AdaIN}(\mathbf{f}^{\vphantom{\mathrm{p}}}_{t|\mathbf{z}}, \mathbf{f}^\mathrm{p}_{0|\mathbf{z}})$, and the source features $\mathbf{f}^\mathrm{p}_{0|\mathbf{z}}$, \ie
\begin{equation}
  \mathcal{L}_\mathrm{con}= \sum_{h=1}^{H} \sum_{w=1}^{W} \sum_{h=c}^{C} \left\lVert \,\mathbf{f}^{\mathrm{p}\ (h,w,c)}_{0|\mathbf{z}}-
  \mathbf{\tilde{f}}^{\vphantom{\mathrm{p}}\ (h,w,c)}_{0|\mathbf{z}}
  \, \right\rVert^2_2 \;,
  \label{eqn:l_cont}
\end{equation}
where $H$, $W$, and $C$ are the height, width, and channels of the feature maps.

\subsection{GAN distillation}
\label{sec:gan_dist}

For continual UDA, in addition to adapting to a new target domain, the generator $G$ shall not \emph{forget} any previously learned knowledge, \ie $G$ should be able to generate data from all previous domains $\mathcal{D}_{0:t-1}$. 
To counteract forgetting, we introduce two distillation losses: $\mathcal{L}^{\mathrm{dis}}_D$ and $\mathcal{L}^{\mathrm{dis}}_G$.
As illustrated in~\Cref{fig:overview}(b), we sample uniformly from the previous domains $\tau \sim \mathcal{U}\{0, t-1\}$ and random noise vectors $\mathbf{z} \sim p_z$, and feed these both to the previous generator $G_\mathrm{p}$ and the current generator $G$.
For distillation, we treat the images generated by the previous generator $\mathbf{x}^\mathrm{p}_{\tau|\mathbf{z}} = G_\mathrm{p}(\tau, \mathbf{z})$ as \emph{real} and the corresponding images generated by the current generator $\mathbf{x}^{\vphantom{\mathrm{p}}}_{\tau|\mathbf{z}} = G(\tau, \mathbf{z})$ as \emph{fake}.
The distillation loss for the discriminator is
\begin{equation}
    \begin{aligned}
    \mathcal{L}^{\mathrm{dis}}_D = \mathop{\mathbb{E}}_{\substack{\mathbf{z} \sim p_z \\ \tau \sim \mathcal{U}\{0,t-1\}}} &\Bigg[ a \bigg( D \Big( \tau, F \big( G(\tau, \mathbf{z}) \big) \Big) \bigg)\\
        &+ a\bigg(\!\!-\!D \Big( \tau, F \big( G_\mathrm{p}(\tau, \mathbf{z}) \big) \Big) \bigg) \Bigg]
    \label{eq:l_dist_d}
    \end{aligned}
\end{equation}
and for the generator
\begin{equation}
    \begin{aligned}
    \mathcal{L}^{\mathrm{dis}}_G = & \overbrace{\!\!\mathop{\mathbb{E}}_{\substack{\mathbf{z} \sim p_z \\ \tau \sim \mathcal{U}\{0,t-1\}}} \Bigg[ a \bigg(\!\!-\!D \Big( \tau, F \big( G(\tau, \mathbf{z}) \big) \Big) \bigg) \Bigg]}^{\mathcal{L}_{G}^\mathrm{adv}} \\
    & + \lambda_\mathrm{img} \underbrace{\!\!\mathop{\mathbb{E}}_{\substack{\mathbf{z} \sim p_z \\ \tau \sim \mathcal{U}\{0,t-1\}}} \Bigg[ \left\Vert G(\tau, \mathbf{z})-G_\mathrm{p}(\tau, \mathbf{z}) \right\Vert_1 \!\Bigg]}_{\mathcal{L}_{G}^\mathrm{img}} \;.
    \end{aligned}
    \label{eq:l_dist_g}
\end{equation}
The first term $\mathcal{L}_{G}^\mathrm{adv}$ in \Cref{eq:l_dist_g} is an adversarial (adv) component evaluated at the feature level, while the second term $\mathcal{L}_{G}^\mathrm{img}$ is an $\ell_1$ image (img) distillation loss at the image level weighted by $\lambda_\mathrm{img}$.
The combination of such image- and feature-level distillation has been empirically demonstrated to be effective for other continual learning tasks, \eg class-incremental learning~\cite{thandiackal2021genifer}.

In practice, GAN adaptation and distillation (\Cref{fig:overview}(a\&b)) are optimized simultaneously by splitting a training batch into two halves: one for adaptation to a new domain, and the other for distillation of previous domains.
This training strategy is more efficient as both the objectives are minimized in the same forward pass.
Then, the final objectives for $D$ and $G$ become
\begin{align}
    \mathcal{L}_D &= \mathcal{L}^\mathrm{uda}_D + \mathcal{L}^\mathrm{dis}_D & \mathrm{and} \qquad \mathcal{L}_G &= \mathcal{L}^\mathrm{uda}_G + \mathcal{L}^\mathrm{dis}_G\;.
    \label{eq:l_g_d}
\end{align}

\subsection{Segmentation training}
\label{sec:seg_training}

Subsequent to adapting the GAN, the segmentation model $M$ is trained to perform segmentation also in the new target domain $\mathcal{D}_t$ (see~\Cref{fig:overview}(c)).
When adapting to $\mathcal{D}_t$, we freeze the previously trained model, called $M_\mathrm{p}$ hereafter, that has seen $\mathcal{D}_{0:t-1}$ and instantiate a trainable model $M$ initialized with the weights of $M_\mathrm{p}$.
$M$ is trained in a supervised manner with images synthesized by $G$ and corresponding pseudo-labels predicted by $M_\mathrm{p}$.
To this end, we employ the concept of knowledge distillation~\cite{Hinton2015Distillation}, \ie the current model $M$ is trained to make the same predictions as the previous model $M_\mathrm{p}$.
Since labeled data is only available during the source training, as described in~\Cref{sec:source_training}, we assume that the predictions of $M_\mathrm{p}$ on the source images are the most accurate compared to images from other target domains.
For knowledge distillation, we first pass a generated source image $\mathbf{x}_{0|\mathbf{z}} = G(0, \mathbf{z})$ to $M_\mathrm{p}$ and obtain the pseudo segmentation map $\hat{m}_0$ at each position $(h,w)$ as
\begin{equation}
    \hat{m}^{(h, w, l)}_0 = \frac{e^{M_\mathrm{p}^{(h, w, l)}\left(\mathbf{x}_{0|\mathbf{z}}\right)/\kappa}}{\sum_{j=1}^{L} e^{M_\mathrm{p}^{(h, w, j)}\left(\mathbf{x}_{0|\mathbf{z}}\right)/\kappa}} \;,
    \label{eq:pseudo_label}
\end{equation}
where $\kappa$ is a temperature parameter that we set to 2, as commonly used in the knowledge distillation literature~\cite{Hinton2015Distillation}.
This produces a softer probability distribution and encodes more fine-grained information compared to one-hot labels.
For the same noise input $\mathbf{z}$, a generated source image $\mathbf{x}_{0|\mathbf{z}}$ and an image $\mathbf{x}_{\tau|\mathbf{z}} = G(\tau, \mathbf{z})$ of any seen domain $\mathcal{D}_{\tau \in \{0, ..., t\}}$ are expected to have the same content, \ie the same ground-truth segmentation map.
We can thus utilize $\hat{m}_0$ as a (silver-standard/approximate) ground truth for $\mathbf{x}_{\tau|\mathbf{z}}$.
Similarly as in~\Cref{eq:l_m_source}, we then train the segmentation model $M$ with the pseudo-labeled samples $(\mathbf{x}_{\tau|\mathbf{z}}, \hat{m}_0)$ by minimizing the cross-entropy loss
\begin{gather}
    \mathcal{L}_M = - \frac{1}{H W} \sum_{h=1}^{H} \sum_{w=1}^{W} \sum_{l=1}^{L} \hat{m}^{(h,w,l)}_0 \log 
    \left( m^{(h,w,l)}_\tau \right) \\
    m^{(h,w,l)}_\tau = \frac{e^{M^{(h, w, l)}\left(\mathbf{x}_{\tau|\mathbf{z}}\right)/\kappa}}{\sum_{j=1}^{L} e^{M^{(h, w, j)}\left(\mathbf{x}_{\tau|\mathbf{z}}\right)/\kappa}}
    \;.
    \label{eq:l_m_target}
\end{gather}

Since the images $\mathbf{x}_{\tau|\mathbf{z}}$ are sampled uniformly from all domains including both the previous ones and the current one, $M$ is trained to generalize across all of them, \ie it adapts to the new target domain $\mathcal{D}_t$ while retaining performance on the previously seen domains $\mathcal{D}_{0:t-1}$.

\section{Experiments}
\label{sec:exp}

We evaluate GarDA on two different tasks, namely optic disc segmentation in CFP images and cardiac segmentation in MR images.
Below, we first describe the employed datasets, preprocessing steps, and implementation details in~\cref{sec:cfp_task,sec:mr_task,sec:implementation}.
Then, we present comparisons to the state of the art in~\cref{sec:sota_results,sec:qual_res}, and discuss further results and ablation studies in~\cref{sec:disc}.

\subsection{Optic disc segmentation (ODS)}
\label{sec:cfp_task}
The task of segmenting the optic disc is challenging and crucial for the clinical detection of glaucoma~\cite{haleem2013automatic}. 
To evaluate our method on this task, we consider three public CFP image datasets, \ie the Retinal Fundus Glaucoma Challenge REFUGE~\cite{orlando2020refuge}, the Indian Diabetic Retinopathy Image Dataset IDRiD~\cite{porwal2018idrid}, and the Retinal Image database for Optic Nerve Evaluation for Deep Learning RIM-ONE\,DL~\cite{batista2020rimone}. 
The datasets were acquired from multiple centers and countries, using different scanners.
Therefore, they allow to create a realistic and representative benchmark for continual UDA.

To construct a continual sequence of domains, we select subsets of the aforementioned datasets where each domain is represented by images acquired with a different camera.
We consider as labeled source domain $\mathcal{D}_0$ the set of images from REFUGE that were acquired with a Canon CR-2 camera.
For the first unlabeled target domain $\mathcal{D}_1$, we use all images from IDRiD, which were captured by a Kowa VX-10$\alpha$ digital fundus camera.
For a second unlabeled target domain $\mathcal{D}_2$, we select the subset of images from RIM-ONE\,DL, which were taken by a Nidek AFC-210 non-mydriatic fundus camera.
The samples from each domain/camera are then randomly split into 80\% training images and 20\% test images, as summarized in~\cref{tab:dat_seq}.
\begin{table}
    \caption{Summary of the datasets used in our continual UDA experiments.}
    \label{tab:dat_seq}
    \centering
    \begin{tabular}{ccccc}
        \toprule
         Task & Domain & \makecell[c]{Device\\vendor} & \makecell[c]{Training\\images} & \makecell[c]{Test\\images} \\
         \midrule
         \multirow{3}{*}{ODS} & $\mathcal{D}_0$ & Canon &  640 & 160 \\
         & $\mathcal{D}_1$ & Kowa &  54 & 27 \\
         & $\mathcal{D}_2$ & Nidek &  278 & 70 \\
         \midrule
         \multirow{3}{*}{CS} & $\mathcal{D}_0$ & Philips &  58 & 16 \\
         & $\mathcal{D}_1$ & Siemens &  76 & 19 \\
         & $\mathcal{D}_2$ & Canon &  40 & 10 \\
        \bottomrule
    \end{tabular}
\end{table}
Note that we only employ a validation set (25\% of the training set) during source segmentation training, since for the training on the target domains only generated images are used.
We follow the preprocessing pipeline employed in~\cite{zhang2019attention} by first locating the optic disc center using a simple boundary detection algorithm~\cite{xu2007optic}. 
Then, the images are cropped to a size of $640 \times 640$ pixels around the estimated optic disc center before being further resized to $256 \times 256$ pixels.

\subsection{Cardiac segmentation (CS)}
\label{sec:mr_task}

We also benchmark our method for CS in MR images, where we segment three cardiac regions: the left ventricle (LV) and the right ventricle (RV) cavities, and the left ventricle myocardium (MYO).
Note that the CS task provides an evaluation setting different from ODS, as it contains higher dynamic-range, grayscale images and aims at multi-class segmentation.

For CS, we use data from the Multi-Centre, Multi-Vendor and Multi-Disease Cardiac Segmentation (M\&Ms) challenge \cite{campello2021mnms}.
It contains volumes from multiple hospitals in Spain and Germany, acquired by 1.5T scanners of different vendors, namely Philips, Siemens, and Canon.
We select the labeled subset of images obtained from the Philips scanner as the source dataset, $\mathcal{D}_0$.
The unlabeled target domains $\mathcal{D}_1$ and $\mathcal{D}_2$ consist of all images acquired using Siemens and Canon scanners, respectively.
Similarly to ODS, we split the data from each domain into 80\% for training and 20\% for testing, where 25\% of the training data is used as validation only during the source segmentation training.
The exact data distribution is shown in~\cref{tab:dat_seq}, where ``image'' denotes a volume, \ie a set of multiple sequential 2D slices.
Note that while our model is trained on 2D slices, the evaluation is run in 3D. 
That is, at test time after 2D inference, the numbers of correctly/falsely classified pixels from all 2D slices in a 3D volume are aggregated to compute a 3D Dice score.

We preprocess the images by employing bias field correction~\cite{tustison2010n4itk} and min-max intensity normalization where the minimum and maximum values are the 5th and 95th intensity percentiles, respectively.
In addition, we resample the 2D slices to an in-plane resolution of $1.2 \times 1.2$~mm, before cropping or padding them to a fixed size of $256 \times 256$ pixels.

\subsection{Implementation details}
\label{sec:implementation}

We use a U-Net~\cite{ronneberger2015unet} segmentation model for GarDA and all the competing methods.
Both the encoder and the decoder consist of five convolutional blocks.
Skip connections are added between each block of the encoder and the corresponding block in the decoder.
We employ the stochastic gradient descent (SGD) optimizer with a learning rate of 0.0005.
Throughout all experimental settings, we use a batch size of 16 and a weight decay of 0.0005.
Source domain training is performed on all available real samples, whereas during the introduction of target domains, arbitrarily many images can be sampled from the generator for training.
Therefore, we measure the training duration for the source domain in number of epochs (\ie passes through the entire dataset) and for adding target domains in number of iterations (\ie batches sampled from $G$).
In particular, the model is trained for 100 epochs on the source domain and for 500 iterations for every target domain addition.

For the generative module, a conditional GAN with a projection discriminator $D$~\cite{Miyato2018CGAN} is used. 
The discriminator incorporates a mini-batch discrimination layer~\cite{karras2017progressive} for better sample diversity.
Note that two types of discriminators are utilized in GarDA.
For the source domain training, the discriminator distinguishes between real and fake images, whereas during adaptation to target adaptation domains, $D$ discriminates features extracted by $F$.
For the source training on images, we use a $D$ architecture with seven convolutional layers.
For target adaptation training, a smaller four-layer backbone is used due to the lower dimensions of the feature space.
To extract these features, we follow previous work~\cite{huang2017adain} and utilize a frozen, pretrained VGG19~\cite{simonyan2014vgg} architecture for $F$.

Our generator $G$ comprises two linear layers followed by nine style-convolution layers~\cite{Karras2020StyleGANv2} consisting of style modulation, convolution, and noise injection.
Besides, we compute an exponential moving average (EMA)~\cite{Yazc2019TheTraining} of $G$'s parameters $\theta$ over the training iterations $n$, \ie $\theta_\mathrm{EMA}^{(n)} = \beta \theta_\mathrm{EMA}^{(n-1)} + (1-\beta) \theta^{(n)}$ with $\theta_\mathrm{EMA}^{(0)} = \theta^{(0)}$ and $\beta=0.999$.
At the end of the training, we construct a copy of $G$ from these averaged parameters, which is afterwards used to generate samples for the training of the segmentation model.
We found such averaging to increase the robustness of $G$ to avoid ending the GAN training in a potentially poor local minimum.
Both $G$ and $D$ are trained with the Adam optimizer and an equalized learning rate~\cite{karras2017progressive} of $0.0025$.
During training, images are augmented with horizontal and vertical flips as well as random rotations (90$^\circ$/180$^\circ$/270$^\circ$).
A summary of the loss weights used for the content loss, image distillation, as well as the $R_1$ regularization in the discriminator is shown in~\cref{tab:hyperparams}.
\begin{table}
    \setlength{\tabcolsep}{3pt}
    \footnotesize
    \caption{Summary of hyperparameter optimization ranges for each method.}
    \label{tab:hyperparams}
    \centering
    \begin{tabular}{llll}
        \toprule
         Method & Parameter & Values & Description \\
         \midrule
         \multirow{3}{*}{GarDA} & $\lambda_\mathrm{con}$ & $\{1, 10\}$ & Content loss \\
         & $\lambda_\mathrm{img}$ & $\{1, 10\}$ & Image distillation \\
         & $\lambda_{R_1}$ & $\{0.2, 2\}$ & $R_1$ regularization \\
         \midrule
         \multirow{2}{*}{\makecell[l]{AdaptSegNet\\\cite{tsai2018adaptsegnet}}} & $\lambda_\mathrm{cp}$ & $\{1, 10, 20, 50\}$ & Class prior loss \\
         & $\lambda_\mathrm{seg}$ & $\{1, 10, 20, 50\}$ & Segmentation loss \\
         \midrule
         \multirow{2}{*}{\makecell[l]{AdvEnt\\\cite{vu2019advent}}} & $\lambda_\mathrm{cp}$ & $\{1, 10, 20, 50\}$ & Class prior loss \\
         & $\lambda_\mathrm{seg}$ & $\{1, 10, 20, 50\}$ & Segmentation loss \\
         \midrule
         \multirow{6}{*}{\makecell[l]{MuHDi\\\cite{saporta2022multi}}} & $\lambda_\mathrm{cp}$ & $\{1, 10, 20, 50\}$ & Class prior loss \\
         & $\lambda_\mathrm{seg}$ & $\{1, 10, 20, 50\}$ & Segmentation loss \\
         & $\lambda_\mathrm{FD}$ & $\{0.01, 0.1, 1\}$ & Feature distillation \\
         & $\lambda_\mathrm{dd}$ & $\{1, 10, 20, 50\}$ & Distribution distillation \\
         & $\lambda_\mathrm{prev}$ & $\{0.1, 0.2, 0.5\}$ & \makecell[lt]{Distribution distillation\\from previous model}\\
         \midrule
         \multirow{3}{*}{\makecell[l]{ACE\\\cite{wu2019ace}}} & $\lambda_\mathrm{style}$ & $\{0.1, 0.5, 1, 5\}$ & Style loss \\
         & $\lambda_\mathrm{cont}$ & $\{0.1, 1\}$ & Content loss \\
         & $\lambda_\mathrm{KL}$ & $\{0.1, 1\}$ & KL divergence loss \\
        \bottomrule
    \end{tabular}
    \vspace{-0.5em}
\end{table}

\subsection{Comparisons with the state of the art}

We compare GarDA to two established non-sequential UDA methods, \ie AdaptSegNet~\cite{tsai2018adaptsegnet} and AdvEnt~\cite{vu2019advent}, and two incremental UDA methods, \ie ACE~\cite{wu2019ace} and MuHDi~\cite{saporta2022multi}.
To enable a fair comparison, we adapt these methods to operate under the continual UDA setting, \ie they do not access real data from the source domain when adapting to the target domains.
Inspired by~\cite{wulfmeier2018incremental}, we therefore use the generator of GarDA that is trained on the source domain, to provide synthesized instead of real source domain images to the above methods.
Apart from this modification, we employ the methods as they were proposed originally.
We implemented ACE from scratch, and the other methods were based on the code provided by MuHDi\footnote{\url{https://github.com/valeoai/MuHDi}}~\cite{saporta2022multi}.
All models were implemented in Pytorch~\cite{Paszke2019PyTorch} and trained on a single NVIDIA A100 GPU.
The hyperparameters for the methods are presented in~\cref{tab:hyperparams}.
Since source-free UDA methods only adapt to a single target domain and DIL methods require labeled target domains, we do not consider such methods for our experimental comparisons.

\subsection{Experimental scenarios}
\label{sec:sota_results}

Segmentation performances were evaluated using the Dice score (aggregated over all foreground labels) achieved \emph{at the end} of the continual domain sequence $\mathcal{D}_0$$\rightarrow$$\mathcal{D}_1$$\rightarrow$$\mathcal{D}_2$, where $\mathcal{D}_0$ denotes the labeled source domain and $\mathcal{D}_1$ \& $\mathcal{D}_2$ represent the unlabeled target domains.
We present the domain-wise Dice scores as well as an overall Dice score aggregated over all domains, which enables the comparison of different methods based on a single number.
The domain-wise scores on the target domains can be regarded as a performance measure for \emph{adaptation}.
Comparisons between GarDA and the competing methods are shown in~\cref{tab:results_dice} for ODS and CS.
\begin{table*}
    \footnotesize
    \setlength{\tabcolsep}{2pt}
    \caption{Domain-wise test Dice scores ($\%$) for optic disc segmentation (ODS) and cardiac segmentation (CS), at the end of the continual sequence $\mathcal{D}_0$\,(Canon)$\rightarrow$ $\mathcal{D}_1$\,(Kowa)$\rightarrow$ $\mathcal{D}_2$\,(Nidek) for ODS, and $\mathcal{D}_0$\,(Philips)$\rightarrow$ $\mathcal{D}_1$~(Siemens)$\rightarrow$ $\mathcal{D}_2$~(Canon) for CS. For the latter, the tabulated results are averages over Dice scores from the end-diastolic (ED) and end-systolic (ES) phases. All results are given as  mean $\pm$ standard deviation (std. dev.) over 3 random initializations.
    The best and the second-best results per column (excluding LB and UB) are  in \textbf{bold} and \underline{underlined}, respectively.}    
    \label{tab:results_dice}
    \centering
    \begin{tabular}{l l cccc c cccc}
        \toprule
        \multirow{2}[1]{*}{Training} & \multirow{2}[1]{*}{Method} & \multicolumn{4}{c}{ODS: Test Dice Scores ($\%$)} & & \multicolumn{4}{c}{CS (ES + ED): Test Dice Scores ($\%$)}\\
        \cmidrule{3-6} \cmidrule{8-11}
        &&$\mathcal{D}_0$ & $\mathcal{D}_1$ & $\mathcal{D}_2$ & overall & & $\mathcal{D}_0$ & $\mathcal{D}_1$ & $\mathcal{D}_2$ & overall\\
        \midrule
        $\mathcal{D}_0$ & LB & $95.5$ & $76.2$ & $44.6$ & $72.1$ & & $84.6$ & $70.9$ & $58.3$ & $71.3$\\
        \midrule
        \multirow{5}{*}{$ \mathcal{D}_0 \rightarrow \mathcal{D}_1 \rightarrow \mathcal{D}_2$} 
         & AdaptSegNet~\cite{tsai2018adaptsegnet} & $\underline{95.4\pm0.0}$ & $72.4\pm3.3$ & $47.3\pm3.1$ & $71.7\pm0.7$ & & $78.8\pm0.7$ & $62.0\pm1.1$ & $48.4\pm1.8$ & $63.1\pm1.2$\\
         & AdvEnt~\cite{vu2019advent} & $\mathbf{95.9\pm0.0}$ & $68.6\pm5.6$ & $52.7\pm3.4$ & $72.3\pm2.9$ & & $\underline{82.2\pm0.1}$ & $67.7\pm1.7$ & $53.8\pm1.3$ & $67.9\pm1.0$\\
         & ACE~\cite{wu2019ace} & $87.8\pm4.1$ & $82.9\pm1.6$ & $57.4\pm4.0$ & $76.0\pm2.6$ & & $82.0\pm0.5$ & $\underline{73.6\pm1.1}$ & $58.6\pm1.0$ & $\underline{71.4\pm0.9}$\\
         & MuHDi~\cite{saporta2022multi} & $92.3\pm3.9$ & $\underline{84.1\pm0.5}$ & $\underline{59.8\pm5.9}$ & $\underline{78.7\pm1.3}$ & & $79.5\pm1.2$ & $71.0\pm2.7$ & $\underline{59.9\pm5.9}$ & $70.1\pm3.3$\\
         & GarDA (ours) & $94.6\pm0.3$ & $\mathbf{87.9\pm0.1}$& $\mathbf{65.2\pm1.5}$ &$\mathbf{82.6\pm0.4}$ & & $\bf{82.8\pm0.1}$ & $\bf{76.0\pm0.2}$ & $\bf{67.2\pm0.1}$ & $\bf{75.3\pm0.1}$\\
        \midrule
        $\mathcal{D}_1$ & SD-UB & $-$ & $92.3$ & $-$ & $-$ & & $-$ & $82.4$ & $-$ & $-$\\
        $\mathcal{D}_2$ & SD-UB & $-$ & $-$ & $93.1$ & $-$ & & $-$ & $-$ & $85.5$ & $-$\\
        $\mathcal{D}_0 + \mathcal{D}_1 + \mathcal{D}_2$ & MD-UB & $96.2$ & $94.7$ & $90.5$ & $93.8$ & & $84.1$ & $83.5$ & $84.5$ & $84.0$\\
        \bottomrule
    \end{tabular}
\end{table*}
For each method, we present the mean and standard deviation (std. dev.) of Dice scores over three different random training initializations.
For reference, we also report lower bound (LB) Dice scores for each domain, which are obtained by training a segmentation model only on the source domain $\mathcal{D}_0$ and then testing on all domains without any adaptation. 
We also report single-domain upper bound (SD-UB) results, \ie the performance of a segmentation model that is trained in a supervised manner (\ie with labels) separately on either $\mathcal{D}_1$ or $\mathcal{D}_2$, and tested on the same corresponding domain.
Finally, the multi-domain upper bound (MD-UB) shows the Dice score achieved by a model that is trained jointly in a supervised fashion using data from all three domains.

\subsubsection{ODS performance analysis}
The results in~\cref{tab:results_dice} show a large gap between LB and SD-UB, \ie 16.1\,pp (percentage points) on $\mathcal{D}_1$ and 48.5\,pp on $\mathcal{D}_2$, indicating that the segmentation model can benefit substantially from adaptation to the target domains.
GarDA is seen to surpass all competing methods in terms of overall Dice score, as well as adaptation performance, \ie domain-wise scores for both target domains.
The non-sequential UDA method AdvEnt achieves the highest Dice score on $\mathcal{D}_0$, but fails to adapt to the target domains.
The same observation can be made for AdaptSegNet.
Among the competing sequential UDA methods, MuHDi and ACE, the former achieves better results on all domains, and overall the second-best in adaptation after our method GarDA.
Our approach outperforms MuHDi's domain-wise mean Dice scores on $\mathcal{D}_0$, $\mathcal{D}_1$, and $\mathcal{D}_2$ by 2.3\,pp, 3.8\,pp, and 5.4\,pp, respectively.
This leads to an improvement of 3.9\,pp in mean overall Dice score -- an improvement of almost 60\% considering that the previous state-of-the-art MuHDi is 6.6\,pp above the no-adaptation LB.
Furthermore, such improvement brings the state of the art in this task by over 25\% closer to its upper bound, given the initial gap of 15.1\,pp from MuHDi to MD-UB.
It is also noteworthy to see that GarDA has by far the smallest standard-deviations for adaptation, which shows the robustness of our method to different initializations -- a very important quality for any deep-learning method and its usability in practice.

\subsubsection{CS performance analysis}
In~\cref{tab:results_dice}, we present the CS results, which are averaged over the Dice scores computed on images from the ED and the ES phases.
Detailed results for the individual ED and ES phases, as well as label-wise Dice scores are reported in Appendix~\ref{app:cs_results}.
Similarly to ODS, CS results in~\cref{tab:results_dice} also show a clear gap between LB and SD-UB, \ie 11.5\,pp on $\mathcal{D}_1$ and 27.2\,pp on $\mathcal{D}_2$.
Compared to ODS, the lower Dice scores for SD-UB suggest the higher complexity of CS, which is plausible since the labels occupy relatively small part of the imaged field and the model aims to segment three instead of one foreground label.
GarDA is seen to outperform all competing methods in all domain-wise and overall Dice scores.
On the source domain, AdvEnt performs comparably to GarDA, but reaches significantly lower adaptation performance in both target domains.
Similarly, AdaptSegNet fails to adapt successfully to $\mathcal{D}_1$ and $\mathcal{D}_2$.
After adapting to the two target domains, both AdvEnt and AdaptSegNet in fact perform inferior to LB.
In comparison to ACE and MuHDi, our method improves domain-wise Dice scores on the target domains $\mathcal{D}_1$ and $\mathcal{D}_2$ by, respectively, 2.4\,pp and 7.3\,pp or more.
Indeed, no competing method provides successful continual domain adaptation, \ie overall Dice scores better than LB, with only ACE yielding 0.1 improvement, but with a standard-deviation of 0.9; whereas GarDA achieves an overall Dice score that is 3.9\,pp higher than ACE's -- the second-best in this task.
Similarly to ODS, the standard deviations of GarDA are also observed to be by far the lowest for any adapted domain and overall, which shows the robustness of this method to initialization.

\subsubsection{Sequential performance analysis}
\label{sec:seq_perf_domain}

\begin{figure}
     \centering
     \begin{subfigure}{0.49\linewidth}
         \centering
         \includegraphics[width=0.98\linewidth]{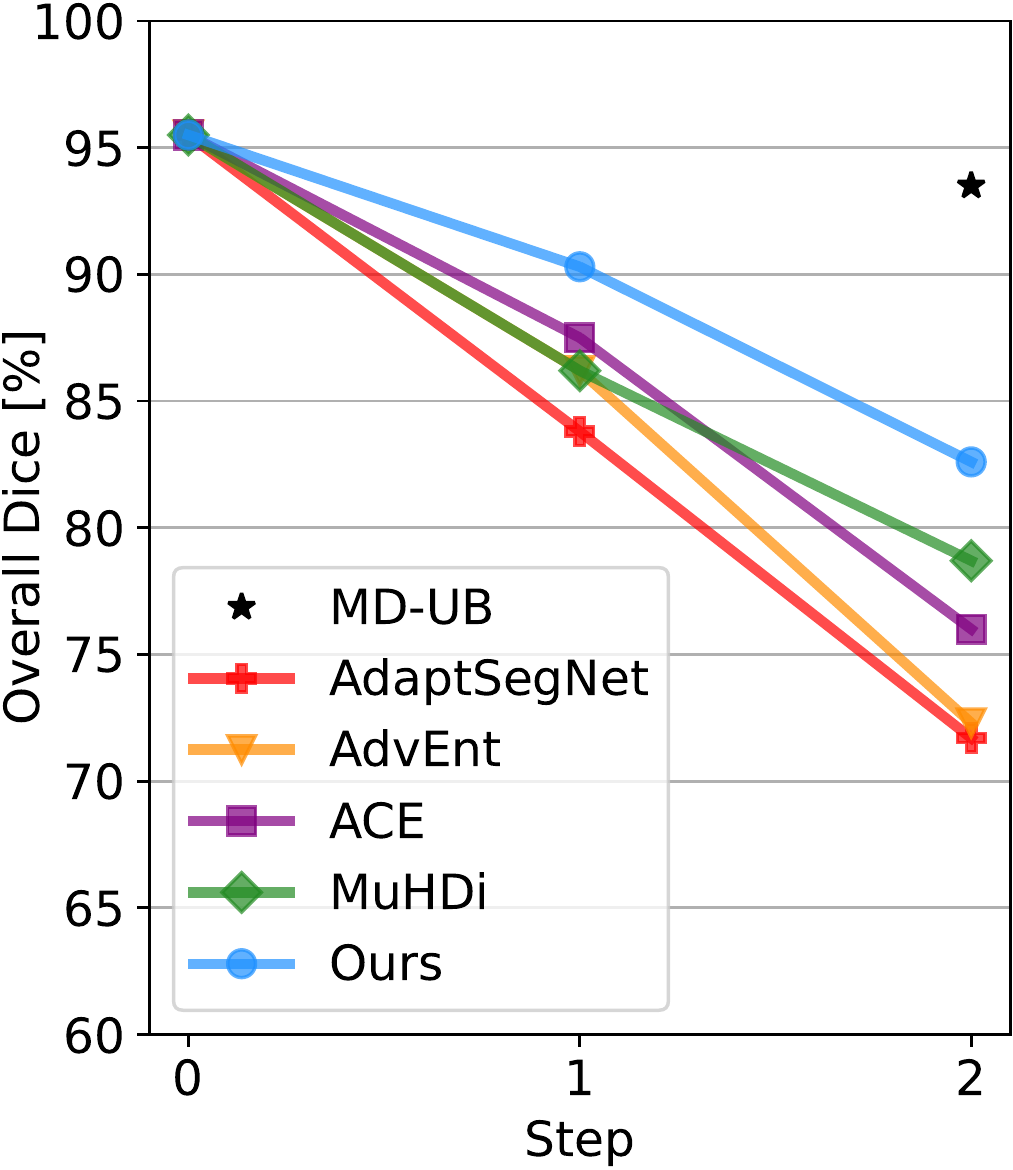}
         \caption{ODS}
     \end{subfigure}
     \begin{subfigure}{0.49\linewidth}
         \centering
         \includegraphics[width=0.98\linewidth]{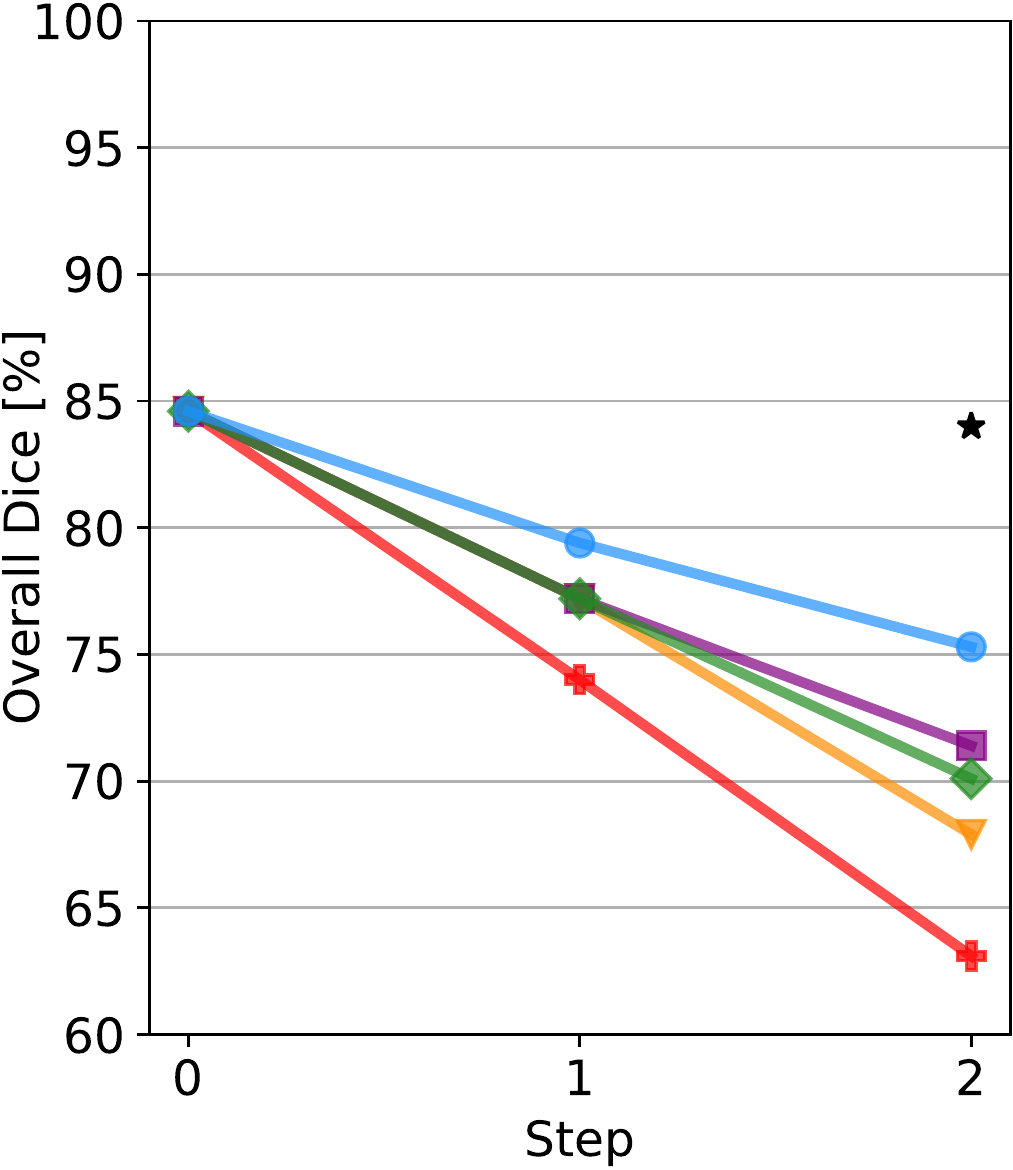}
         \caption{CS}
     \end{subfigure}
     \caption{Evolution of the mean overall Dice score over the seen domains after each training step in the continual sequence for (a) ODS and (b) CS.}
     \label{fig:avg_dice_evol}
\end{figure}

In addition to the Dice scores achieved at the end of the domain sequence, we also analyze the evolution of overall segmentation performance while the model adapts to each domain in~\cref{fig:avg_dice_evol}.
At each step of the sequence, the overall Dice score is computed over all domains seen up to that step.
As seen in~\cref{fig:avg_dice_evol}, the overall scores decrease during the sequence as the models need to address an increasing number of domains.
It can be observed that the non-sequential UDA baselines AdaptSegNet and AdvEnt exhibit the largest drop in overall score compared to the initial performance on $\mathcal{D}_0$.
For ODS, both methods perform already subpar compared to the other approaches after adapting to $\mathcal{D}_1$, whereas for CS, AdaptSegNet still achieves a similar overall score as ACE and our approach GarDA.
However, both AdaptSegNet and AdvEnt fail to adapt effectively to $\mathcal{D}_2$ in the last step, which causes their final overall score to be the lowest among all methods.
Note that since MuHDi is a continual extension of AdvEnt, their performances up until $\mathcal{D}_1$ are identical.
The benefits of MuHDi over AdvEnt in a continual UDA setting are only seen in the last step, by the gain in overall Dice score, \ie 6.4\,pp for ODS and 2.2\,pp for CS.
MuHDi outperforms ACE on ODS, but for CS, attains slightly lower overall scores at the end of the sequence.
In comparison, our proposed method GarDA substantially outperforms all other methods throughout the entire domain sequence for both ODS and CS.

\subsection{Qualitative results}
\label{sec:qual_res}

\begin{figure*}
  \centering
  \includegraphics[width=0.75\linewidth]{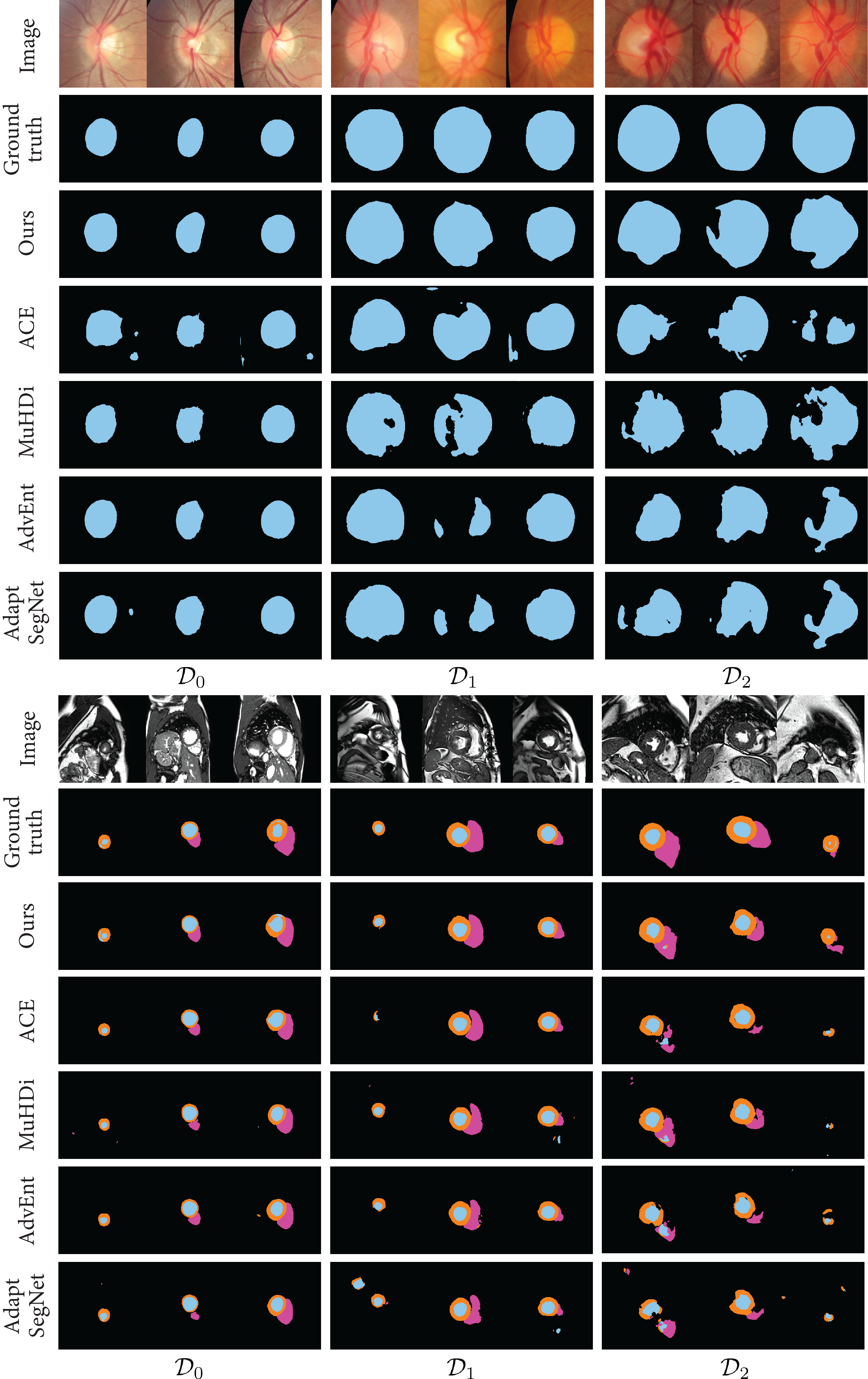}
  \caption{Examples of real test images and our corresponding predicted segmentation maps. For ODS (top), the predicted optic disc is colored in blue. For CS (bottom), the left ventricle (LV) is colored in orange, the right ventricle (RV) in purple, and the myocardium (MYO) in blue.}
  \label{fig:qual_res}
\end{figure*}

In~\cref{fig:qual_res}, we present test images for ODS and CS along with the corresponding segmentation maps predicted by GarDA and the competing methods.
It can be observed that GarDA is able to accurately segment the optic disc as well as cardiac regions across different domains.
As already highlighted in the quantitative results, our method performs particularly better on the last target domain in comparison to the other methods.
Although not perfect, GarDA is able to segment most of the anatomical structures, whereas the other methods miss many of the relevant parts; highlighted 
specifically by the results in the last columns of $\mathcal{D}_2$.

\begin{figure*}
  \centering
  \includegraphics[width=0.9\linewidth]{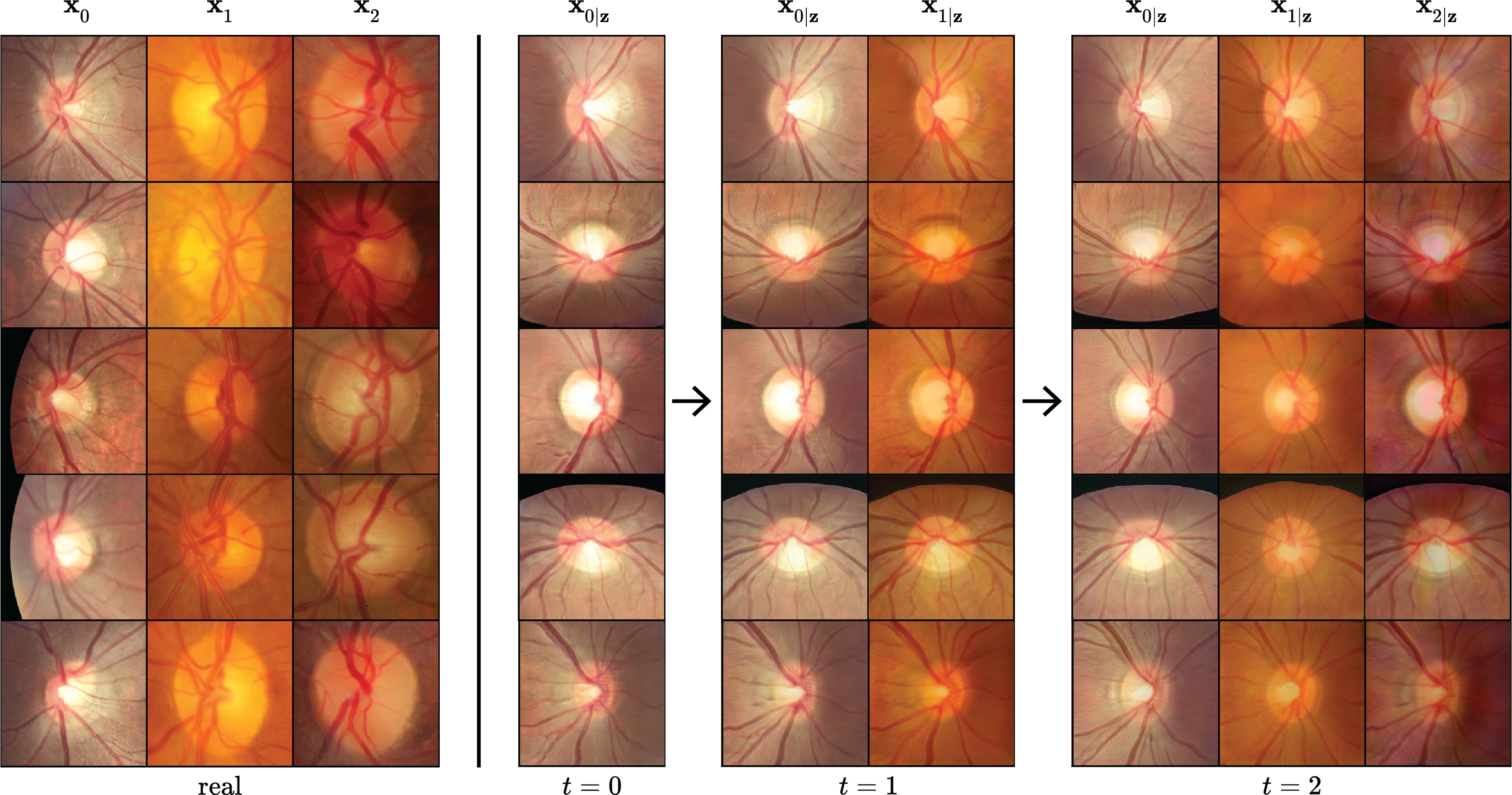}
  \caption{Real sample images (left) from the ODS dataset domains, and (right) those sampled from GarDA's generator at steps $t=0,1,2$.}
  \label{fig:generated_ods}
\end{figure*}

To give an insight into how our GAN works, in~\cref{fig:generated_ods} we show examples of real images and images synthesized by our generator after each adaptation step in the continual domain sequence.
It can be observed that the generator successfully learns to create images with the appearance characteristics of each domain.
In particular, our generator captures the orange tint of $\mathcal{D}_1$ as well as the darker appearance of $\mathcal{D}_2$ with the more pronounced vessels in dark-red colors.
Furthermore, \cref{fig:generated_ods} shows that the content in the synthesized images stays relatively consistent across domains, which is crucial for the segmentation training to be effective.
We can also see at step $t=2$ that, thanks to the GAN distillation losses, the generated images of previously seen domains do not change significantly compared to $t=0$ or $t=1$, when the domains were first learned.
Note that some of the generated images appear to be rotated. It can be attributed to the employment of flipping and random rotations during GAN training, which causes the generator to reproduce such augmentations.
However, this is not an issue since these images are merely used to train the segmentation model, which can accommodate any orientation.

\begin{figure*}
  \centering
  \includegraphics[width=0.9\linewidth]{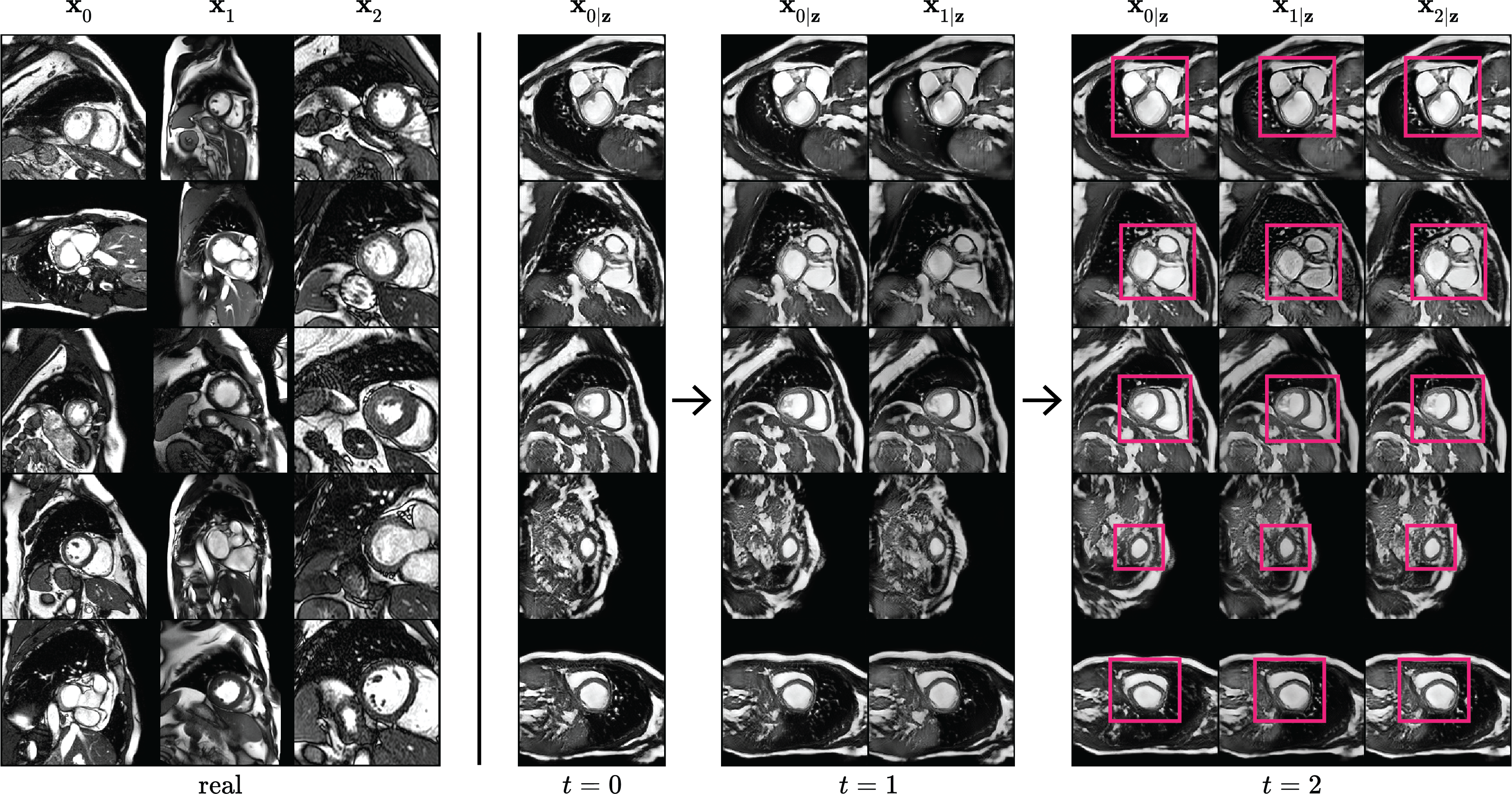}
  \caption{Real sample images (left) from the CS dataset domains, and (right) those sampled from GarDA's generator at steps $t=0,1,2$.
  The pink rectangles serve as visual aid for the comparison of relevant anatomical structures between different domains.}
  \label{fig:generated_cs}
\end{figure*}

Analogously, we show in~\cref{fig:generated_cs} our generated images for CS after each step in the continual domain sequence.
Therefore, we include reference rectangles to better emphasize appearance differences between the domains around the anatomy of interest for segmentation.
In particular, images from $\mathcal{D}_1$ tend to be darker and sometimes blurry, while images from $\mathcal{D}_0$ appear to be the brightest overall.
As MR images are in grayscale and with high-dynamic range, the differences between the domains are not as evident to the human eye as in the case of retinal RGB images.
Nevertheless, the results we presented earlier indicate that such differences still significantly hinder generalization of models between domains, thereby necessitating effective continual adaptation schemes. 
Similarly to ODS, our generator is able to create CS images with different domain characteristics for the same anatomical content.
Additionally, GarDA's generator does not suffer from catastrophic forgetting of previously seen domains.

\subsection{Discussion}
\label{sec:disc}
    
In the following subsections, we provide more insights into our proposed method and the presented results.
In particular, we analyze the benefits of continual UDA over single-step UDA in~\cref{sec:fwd_bwd_transfer} and discuss the advantages of GarDA over ACE in~\cref{sec:comp_ace}, which also employs a form of appearance transfer.
Finally, we examine the effectiveness of each of our proposed contributions via ablation studies in~\cref{sec:ablations}.

\subsubsection{Forward and backward transfer}
\label{sec:fwd_bwd_transfer}

\begin{figure}
  \centering
  \begin{subfigure}{0.49\linewidth}
     \centering
     \includegraphics[width=0.98\linewidth]{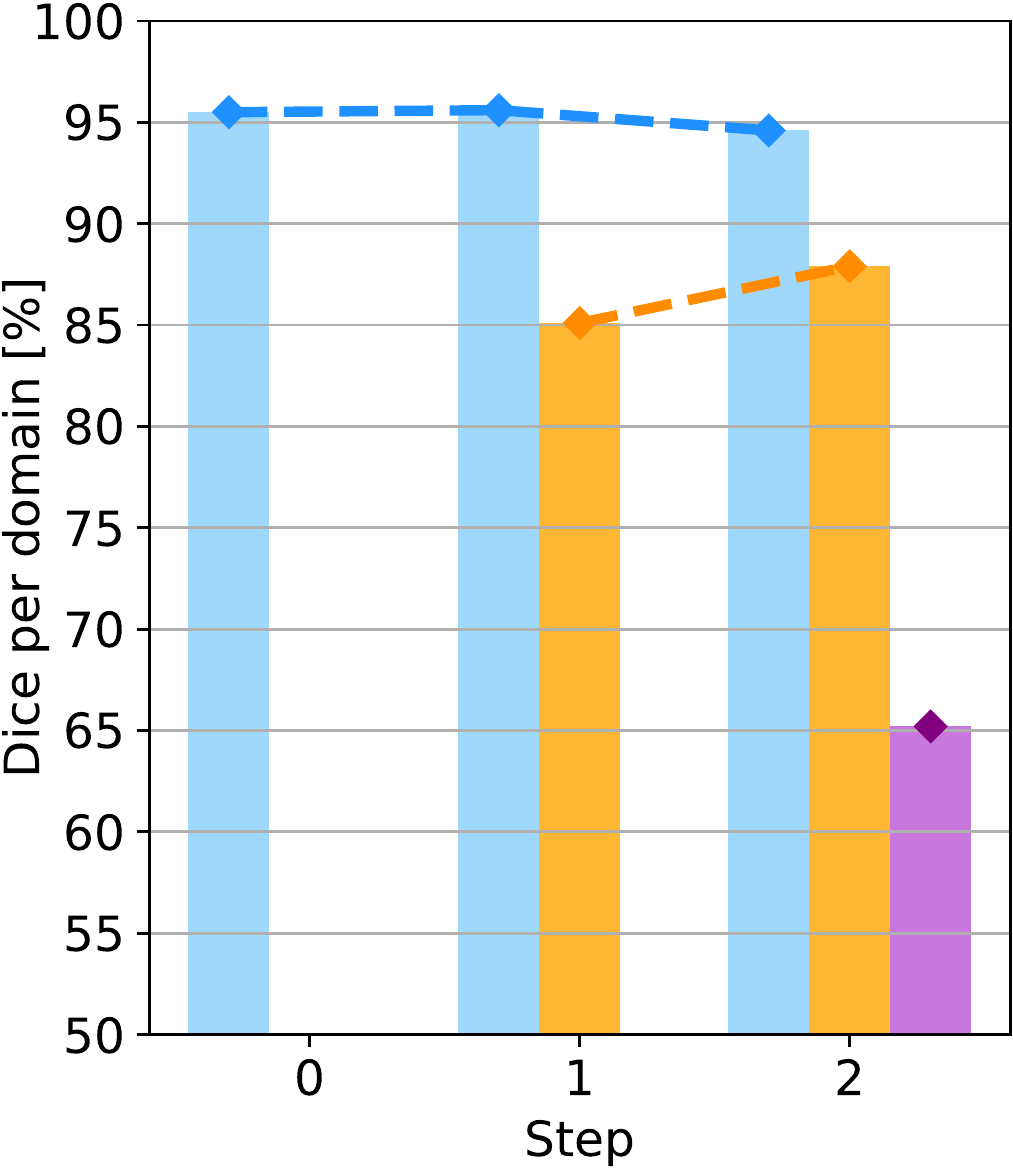}
     \caption{ODS}
 \end{subfigure}
 \begin{subfigure}{0.49\linewidth}
     \centering
     \includegraphics[width=0.98\linewidth]{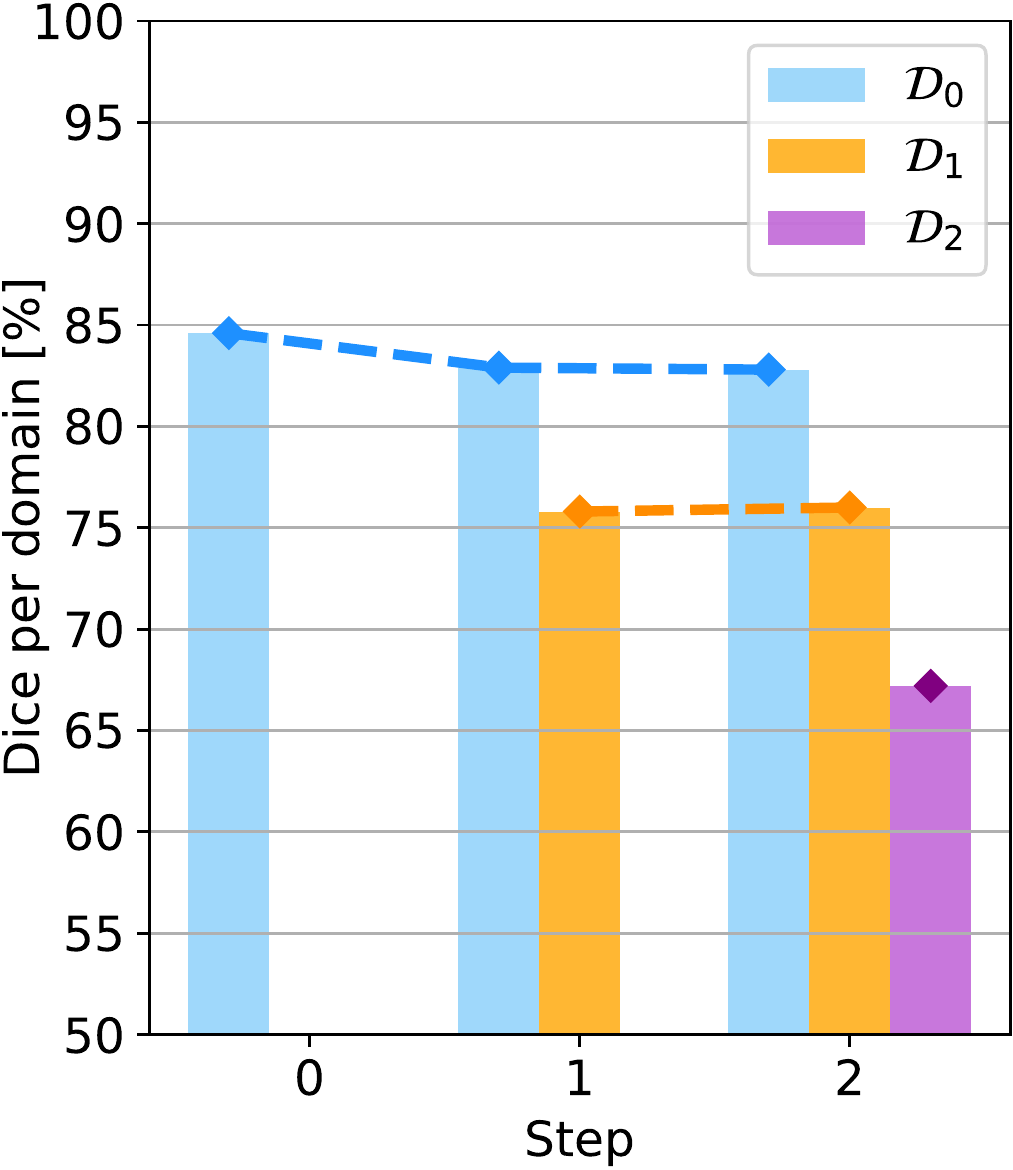}
     \caption{CS}
 \end{subfigure}
  \caption{Evolution of the mean domain-wise Dice scores on $\mathcal{D}_0$, $\mathcal{D}_1$, and $\mathcal{D}_2$ after each step of training in the continual sequence for (a) ODS and (b) CS.}
  \label{fig:domain_dice_evolution}
\end{figure}

An alternative to continually adapting to new domains would be to only perform single-step UDA, \ie $\mathcal{D}_0$$\rightarrow$$\mathcal{D}_1$ or $\mathcal{D}_0$$\rightarrow$$\mathcal{D}_2$.
We show in~\cref{tab:transfer}, that these alternatives are in fact inferior to the model that is continually trained on all domains $\mathcal{D}_0$$\rightarrow$$\mathcal{D}_1$$\rightarrow$$\mathcal{D}_2$.
\begin{table}
    \notsotiny
    \setlength{\tabcolsep}{1pt}
    \caption{Test Dice scores ($\%$) for optic disc segmentation (ODS) and cardiac segmentation (CS) on (a)~$\mathcal{D}_1$ and (b)~$\mathcal{D}_2$. Results for CS are averaged over the end-diastolic (ED) and end-systolic (ES) phases. Compared models are either adapted to only a single domain (1st row) or continually adapted to both target domains (2nd row). All results show the mean $\pm$ std-dev over 3 random initializations.
    Superior results for the column are shown in \textbf{bold}.}
    \label{tab:transfer}
    \begin{subtable}{0.49\linewidth}
        \centering
        \caption{$\mathcal{D}_1$}
        \label{tab:transfer_d1}
        \begin{tabular}{lcc}
        \toprule
        \multirow{2}[1]{*}{Training} & \multicolumn{2}{c}{Test Dice Scores ($\%$)} on $\mathcal{D}_1$\\
        \cmidrule{2-3}
        & ODS & CS (ES + ED) \\
        \midrule
        $\mathcal{D}_0$$\rightarrow$$\mathcal{D}_1$ & $85.1\pm0.2$ & $75.8\pm0.0$ \\
        $\mathcal{D}_0$$\rightarrow$$\mathcal{D}_1$$\rightarrow$$\mathcal{D}_2$ & $\mathbf{87.9\pm0.1}$ & $\mathbf{76.0\pm0.2}$ \\
        \bottomrule
        \end{tabular}
    \end{subtable}
    \begin{subtable}{0.49\linewidth}
        \centering
        \caption{$\mathcal{D}_2$}
        \label{tab:transfer_d2}
        \begin{tabular}{lcc}
        \toprule
        \multirow{2}[1]{*}{Training} & \multicolumn{2}{c}{Test Dice Scores ($\%$)} on $\mathcal{D}_2$\\
        \cmidrule{2-3}
        & ODS & CS (ES + ED) \\
        \midrule
        $\mathcal{D}_0$$\rightarrow$$\mathcal{D}_2$ & $63.9\pm1.2$ & $63.3\pm0.2$ \\
        $\mathcal{D}_0$$\rightarrow$$\mathcal{D}_1$$\rightarrow$$\mathcal{D}_2$ & $\mathbf{65.2\pm1.5}$ & $\mathbf{67.2\pm0.1}$ \\
        \bottomrule
        \end{tabular}
    \end{subtable}
\end{table}
For both ODS and CS, the continually trained model achieves better performance on $\mathcal{D}_1$ compared to UDA directly to $\mathcal{D}_1$, which suggests that the adaptation to a later target domain ($\mathcal{D}_2$) can indeed provide useful information and improved performance for a previous target domain.
This can also be observed in the evolution of the domain-wise Dice scores in~\cref{fig:domain_dice_evolution}, especially for ODS.
In the continual learning literature, this phenomenon is known as \emph{backward transfer}~\cite{lopez2017gem} of learned information.
In addition, we see in~\cref{tab:transfer_d2} that the intermediate adaptation to $\mathcal{D}_1$ helps to improve the final Dice score on $\mathcal{D}_2$ for both ODS and CS.
This is known as \emph{forward transfer}~\cite{lopez2017gem}.
Both transfers indicate that, when effectively incorporated, information from additional domains can improve the accuracy and generalizability of a model also on other domains.
Owing to this, we find continual UDA to overall perform superior to single-step UDA.
Furthermore, the continually trained model is able to segment images from all involved domains, not only two.
This is generally beneficial, as one needs to keep only a single model with accumulated information.
For instance, when a care facility acquires a new imaging device, with single-step UDA, they would need to find the right model suitable for this device (or collect new data to adapt from source) but with continual UDA, they can obtain the latest continual model and expect it to perform well out-of-the-box if it has seen similar images earlier.

\subsubsection{Comparison to ACE}
\label{sec:comp_ace}

\begin{figure*}
  \centering
  \includegraphics[width=\linewidth]{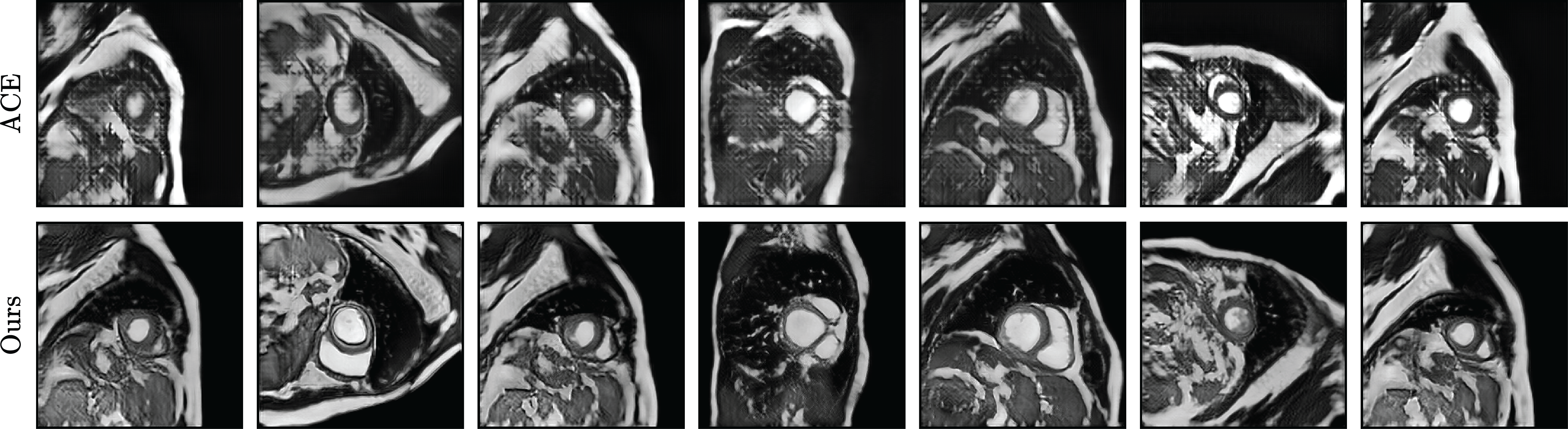}
  \caption{Examples of cardiac magnetic resonance (MR) images synthesized given the same content $\mathbf{z}$ by ACE (top) and our proposed generator (bottom).}
  \label{fig:ace_vs_ours}
\end{figure*}

In order to generate images from previous target domains, ACE first extracts feature maps from a source image and re-normalizes them with appearance statistics drawn from a memory buffer.
The memory buffer represents previously seen target domain images in the form of mean and variance values of feature maps.
The re-normalized features are then fed to a decoder (an inverted VGG19~\cite{simonyan2014vgg}) that maps the features to the desired target images.
There are several shortcomings with this approach:
First, ACE stores a fixed number of mean and variance values in the memory buffer, which cannot accurately represent the appearance information of the entire distribution of target images.
In contrast, we employ a GAN trained to reproduce images from this entire distribution.
Indeed we illustrate in~\cref{fig:ace_vs_ours} the superior quality of the images generated by our approach.
While ACE's synthesized images exhibit severe checkerboard artifacts, the images produced by our generator appear cleaner and closer to real samples.
Second, ACE always requires a source image and a set of target domain statistics to generate a target image, which limits the number and diversity of samples that can be generated.
On the other hand, our proposed generator is able to produce arbitrarily many images from any domain using a simple random noise vector input and a domain label.
Third, the inverted VGG decoder of ACE comprises more than 140 million parameters, whereas our GAN (generator and discriminator together) contains merely 23 million parameters.
Thus, the performance gain from our method GarDA also comes at a lower budget.

\subsubsection{Ablation study}
\label{sec:ablations}

We conduct ablation experiments for each proposed component in GarDA to quantify their individual impact on the final segmentation performance.
This evaluation is performed for ODS since the large gap between lower and upper bound allows us to clearly demonstrate performance gains.
All results are summarized in~\cref{tab:ablation}.
\begin{table}
    \notsotiny
    \setlength{\tabcolsep}{1pt}
    \def\arraystretch{1.4}
    \caption[Ablation]{Test Dice scores ($\%$) of ablations on optic disc segmentation (ODS), at the end of the continual sequence $\mathcal{D}_0$ (Canon) $\rightarrow$$\mathcal{D}_1$ (Kowa)  $\rightarrow$$\mathcal{D}_2$ (Nidek). For clarity, mean results are shown without std. dev. The difference between our method and its variations with ablated components is shown in parentheses.}
    \label{tab:ablation}
    \centering
    \begin{tabular}{lllllll}
        \toprule
        \multirow{2}{*}{Training} & \multirow{2}{*}{Method} & \multicolumn{4}{c}{Test Dice Scores ($\%$)} \\
        \cmidrule{3-6}
        &&$\mathcal{D}_0$ & $\mathcal{D}_1$ & $\mathcal{D}_2$ & overall \\
        \midrule
        $\mathcal{D}_0$ & LB & $95.5$ & $76.2$ & $44.6$ & $72.1$ \\
        \midrule
        \multirow{4}{*}{$ \mathcal{D}_0$$\rightarrow$$\mathcal{D}_1$$\rightarrow$$\mathcal{D}_2$}
        & Ours\,$-$\,$\mathcal{L}_\mathrm{con}$ & $85.9$ \diff{-8.8} & $82.9$ \diff{-5.0} & $72.0$ \textcolor{blue}{(+6.8)} & $80.3$ \diff{-2.3} \\
        & Ours\,$-$\,$\mathcal{L}_{G}^\mathrm{img}$ & $82.2$ \diff{-12.5} & $77.1$ \diff{-10.8} & $52.9$ \diff{-12.3} & $70.7$ \diff{-11.9} \\
         & Ours\,$-$\,$\mathcal{L}^\mathrm{adv}$ & $87.5$ \diff{-7.1} & $76.8$ \diff{-11.1} & $53.2$ \diff{-12.1} & $72.5$ \diff{-10.1} \\
         & Ours & $94.6$ & $87.9$ & $65.2$ & $82.6$\\
        \midrule
        $\mathcal{D}_1$ & SD-UB & $-$ & $92.3$ & $-$ & $-$ \\
        $\mathcal{D}_2$ & SD-UB & $-$ & $-$ & $93.1$ & $-$ \\
        $\mathcal{D}_0$$+$$\mathcal{D}_1$$+$$\mathcal{D}_2$ & MD-UB & $96.2$ & $94.7$ & $90.5$ & $93.8$ \\
        \bottomrule
    \end{tabular}
\end{table}

The impact of the content loss can be seen in comparison to the ablation ``Ours$-$$\mathcal{L}_\mathrm{con}$''.
Omitting the content loss causes the model to lose performance on previously seen domains, exemplified by drops in domain-wise Dice scores by 8.8\,pp and 5.0\,pp on $\mathcal{D}_0$ and $\mathcal{D}_1$, respectively.
Interestingly, the removal of the content loss increases the Dice score on the last target domain by 6.8\,pp.
This is likely because the generator is less constrained to retain the same content across domains, which allows it to better reproduce the appearance of the newer (last) domain.
However, this then comes at the cost of a larger drop in performance on the earlier domains.

Our proposed GAN distillation losses, consisting of $\mathcal{L}_G^\mathrm{dis}$$=$$\mathcal{L}_{G}^\mathrm{adv}$$+$$\lambda_\mathrm{img}\mathcal{L}_{G}^\mathrm{img}$ for the generator~(\cref{eq:l_dist_g}) and $\mathcal{L}_D^\mathrm{dis}$ for the discriminator~(\cref{eq:l_dist_d}), are designed to prevent forgetting of the source and intermediate target domains.
We analyze the individual effects of adversarial and non-adversarial components.
The only non-adversarial component is the image distillation loss, the ablation of which we tabulate as ``Ours$-$$\mathcal{L}_{G}^\mathrm{img}$''.
The results show that omitting image distillation has a major -- indeed the biggest -- impact on the segmentation performance.
Namely, the Dice scores drop for all domains by more than 10.0\,pp.
For ablating the adversarial components $\mathcal{L}^\mathrm{adv}$$=$$\mathcal{L}_{G}^\mathrm{adv}$$+$$\mathcal{L}_D^\mathrm{dis}$, we tabulate the results as ``Ours$-$$\mathcal{L}^\mathrm{adv}$''.
The impact is similar to the image distillation, but does not affect the source domain performance as much.
In summary, our ablation study confirms that each of the proposed components is essential to achieve our new state-of-the-art segmentation performance for continual domain adaptation.

\section{Conclusions}
\label{sec:conclusions}

In this work, we have proposed GarDA, a novel segmentation method for continual UDA that can adapt to new domains without forgetting previous ones.
As GarDA employs generative appearance replay, it does not require to store previously seen data, neither from the source domain nor from any intermediate target domains.
To the best of our knowledge, GarDA is the first segmentation method for such strictly continual UDA.
This makes our approach widely applicable in practice, where data from previous domains cannot be retained indefinitely.
A potential limitation of our approach is the duration of the GAN training, which may take several hours depending on the desired quality of the domain appearance adaptation.
However, since new data from different domains usually does not become available on a daily basis, this should not be an issue in practice.

We evaluate the generalizability of our method by conducting comprehensive experiments for two very different tasks, \ie optic disc segmentation on color fundus photography images and cardiac segmentation on magnetic resonance images.
Our results demonstrate that GarDA substantially outperforms all competing methods on both tasks.
In addition, we provide qualitative results showing that our proposed generator is able to synthesize meaningful images from different domains that enable the segmentation model to be trained continually.
Our ablation study highlights the impact of each proposed major component in GarDA, \ie the content loss, image distillation, and adversarial distillation.
Overall, the trends observed in our experiments indicate GarDA to potentially yield better results for increasingly longer domain sequences.
In the future, when new datasets from different domains become available, longer continual sequences with more domains shall be studied.

\section*{Acknowledgments}
We would like to thank Krishna Chaitanya and Neerav Karani for insightful discussions as well as their valuable advice on dataset selection and preprocessing.

\newpage
\bibliographystyle{IEEEtran}
\bibliography{refs}

\appendices
\renewcommand{\thetable}{A\arabic{table}}
\setcounter{table}{0}

\section{Detailed cardiac segmentation results}
\label{app:cs_results}

Individual results for the end-systolic (ES) and end-diastolic (ED) phases are presented in~\cref{tab:results_cs_es} and \cref{tab:results_cs_ed}.
For the majority of labels and domains, it can be observed that our approach achieves the best or the second-best results in terms of label-wise and domain-wise Dice scores.
\begin{sidewaystable}
    \scriptsize
    \setlength{\tabcolsep}{2pt}
    \caption{Label-wise test Dice scores ($\%$) for cardiac segmentation (CS) in the end-systolic (ES) phase, at the end of the continual sequence $\mathcal{D}_0$ (Philips)$\rightarrow$$\mathcal{D}_1$ (Siemens)$\rightarrow$$\mathcal{D}_2$ (Canon).
    All results are given as  mean $\pm$ std. dev. over 3 random initializations.
    The best and the second-best results per column (excluding LB and UB) are in \textbf{bold} and \underline{underlined}, respectively.}
    \label{tab:results_cs_es}
    \centering
    \begin{tabular}{llccccccccccccc}
        \toprule
        \multirow{3}[3]{*}{Training} & \multirow{3}[3]{*}{Method} & \multicolumn{13}{c}{Test Dice Scores ES ($\%$)} \\
        \cmidrule{3-15}
        & & \multicolumn{4}{c}{$\mathcal{D}_0$} & \multicolumn{4}{c}{$\mathcal{D}_1$} & \multicolumn{4}{c}{$\mathcal{D}_2$} & \multirow{2}[1]{*}{overall} \\
        \cmidrule(lr){3-6} \cmidrule(lr){7-10} \cmidrule(lr){11-14}
        & & MYO & LV & RV & avg & MYO & LV & RV & avg & MYO & LV & RV & avg \\
        \midrule
        $\mathcal{D}_0$ & LB & $81.0$ & $82.0$ & $80.5$ & $81.2$ & $76.3$ & $68.1$ & $64.0$ & $69.5$ & $69.6$ & $66.9$ & $45.0$ & $60.5$ & $70.4$ \\
        \midrule
        \multirow{5}{*}{$\mathcal{D}_0$$\rightarrow$$\mathcal{D}_1$$\rightarrow \mathcal{D}_2$}
        & AdaptSegNet~\cite{tsai2018adaptsegnet} & $76.2\pm1.0$ & $77.7\pm1.0$ & $70.2\pm0.7$ & $74.7\pm0.8$ & $72.2\pm0.6$ & $58.5\pm1.2$ & $52.8\pm1.5$ & $61.2\pm1.0$ & $55.7\pm3.1$ & $52.1\pm2.2$ & $35.1\pm1.6$ & $47.7\pm2.2$ & $61.2\pm1.1$ \\
        & AdvEnt~\cite{vu2019advent} & $\bf{80.0\pm0.7}$ & $\bf{81.0\pm0.5}$ & $74.3\pm1.5$ & $78.4\pm0.5$ & $\underline{78.0\pm0.7}$ & $65.3\pm1.3$ & $55.3\pm2.5$ & $66.2\pm1.1$ & $66.8\pm1.5$ & $58.8\pm2.1$ & $35.9\pm2.6$ & $53.8\pm0.9$ & $66.1\pm0.8$ \\
        & ACE~\cite{wu2019ace} & $\underline{79.5\pm0.2}$ & $79.1\pm1.5$ & $\underline{76.8\pm0.2}$ & $\underline{78.5\pm0.6}$ & $\bf{79.9\pm0.3}$ & $\underline{68.8\pm0.7}$ & $\underline{65.9\pm1.6}$ & $\underline{71.6\pm0.8}$ & $\bf{71.0\pm0.4}$ & $\underline{64.8\pm1.2}$ & $41.6\pm1.8$ & $\underline{59.1\pm1.0}$ & $\underline{69.7\pm0.4}$ \\
        & MuHDi~\cite{saporta2022multi} & $74.5\pm2.0$ & $76.8\pm1.3$ & $73.1\pm5.2$ & $74.8\pm2.0$ & $77.8\pm1.9$ & $64.7\pm3.2$ & $60.6\pm4.0$ & $67.7\pm2.8$ & $66.1\pm3.0$ & $56.1\pm7.1$ & $\underline{44.7\pm5.9}$ & $55.6\pm5.1$ & $62.4\pm2.7$ \\
        & GarDA (ours) & $78.0\pm0.1$ & $\underline{79.5\pm0.2}$ & $\bf{81.3\pm0.0}$ & $\bf{79.6\pm0.1}$ & $77.2\pm0.3$ & $\bf{72.0\pm0.2}$ & $\bf{73.4\pm0.1}$ & $\bf{74.2\pm0.2}$ & $\underline{69.6\pm0.2}$ & $\bf{72.5\pm0.1}$ & $\bf{60.4\pm0.2}$ & $\bf{67.5\pm0.1}$ & $\bf{73.8\pm0.1}$ \\
        \midrule
        $\mathcal{D}_1$ & SD-UB & $-$ & $-$ & $-$ & $-$ & $83.1$ & $76.7$ & $79.4$ & $79.7$ & $-$ & $-$ & $-$ & $-$ & $-$\\
        $\mathcal{D}_2$ & SD-UB & $-$ & $-$ & $-$ & $-$ & $-$ & $-$ & $-$ & $-$ & $82.9$ & $84.5$ & $80.6$ & $82.7$ & $-$\\
        $\mathcal{D}_0$$+$$\mathcal{D}_1$$+$$\mathcal{D}_2$ & MD-UB & $79.2$ & $79.7$ & $78.7$ & $79.2$ & $85.9$ & $76.3$ & $78.4$ & $80.2$ & $83.8$ & $82.9$ & $77.4$ & $81.3$ & $80.2$ \\
        \bottomrule
    \end{tabular}
\end{sidewaystable}

\begin{sidewaystable}
    \scriptsize
    \setlength{\tabcolsep}{2pt}
    \caption{Label-wise test Dice scores ($\%$) for cardiac segmentation (CS) in the end-diastolic (ED) phase, at the end of the continual sequence $\mathcal{D}_0$ (Philips)$\rightarrow$$\mathcal{D}_1$ (Siemens)$\rightarrow$$\mathcal{D}_2$ (Canon). All results are given as  mean $\pm$ std. dev. over 3 random initializations.
    The best and the second-best results per column (excluding LB and UB) are in \textbf{bold} and \underline{underlined}, respectively.}
    \label{tab:results_cs_ed}
    \centering
    \begin{tabular}{llccccccccccccc}
        \toprule
        \multirow{3}[3]{*}{Training} & \multirow{3}[3]{*}{Method} & \multicolumn{13}{c}{Test Dice Scores ED ($\%$)} \\
        \cmidrule{3-15}
        & & \multicolumn{4}{c}{$\mathcal{D}_0$} & \multicolumn{4}{c}{$\mathcal{D}_1$} & \multicolumn{4}{c}{$\mathcal{D}_2$} & \multirow{2}[1]{*}{overall} \\
        \cmidrule(lr){3-6} \cmidrule(lr){7-10} \cmidrule(lr){11-14}
        & & MYO & LV & RV & avg & MYO & LV & RV & avg & MYO & LV & RV & avg \\
        \midrule
        $\mathcal{D}_0$ & LB & $94.5$ & $82.9$ & $86.9$ & $88.1$ & $84.5$ & $67.6$ & $64.6$ & $72.2$ & $77.7$ & $53.8$ & $37.0$ & $56.2$ & $72.2$ \\
        \midrule
        \multirow{5}{*}{$\mathcal{D}_0 \rightarrow \mathcal{D}_1 \rightarrow \mathcal{D}_2$}
        & AdaptSegNet~\cite{tsai2018adaptsegnet} & $91.0\pm0.7$ & $78.0\pm0.9$ & $80.1\pm0.6$ & $83.0\pm0.6$ & $79.4\pm1.2$ & $53.9\pm1.3$ & $55.3\pm2.1$ & $62.9\pm1.2$ & $64.3\pm2.3$ & $44.6\pm0.8$ & $38.6\pm2.2$ & $49.2\pm1.3$ & $65.0\pm0.9$ \\
        & AdvEnt~\cite{vu2019advent} & $\bf{93.3\pm0.7}$ & $\bf{81.4\pm0.9}$ & $83.2\pm1.0$ & $\underline{86.0\pm0.3}$ & $85.1\pm0.4$ & $63.5\pm1.7$ & $58.7\pm5.1$ & $69.1\pm2.1$ & $72.9\pm3.2$ & $52.2\pm7.1$ & $36.2\pm7.8$ & $53.7\pm1.7$ & $69.6\pm1.2$ \\
        & ACE~\cite{wu2019ace} & $\underline{93.1\pm0.4}$ & $79.7\pm1.0$ & $\underline{83.8\pm0.1}$ & $85.5\pm0.4$ & $\bf{87.4}\pm0.5$ & $\underline{69.0\pm1.3}$ & $70.4\pm2.2$ & $\underline{75.6\pm1.3}$ & $78.6\pm0.5$ & $\underline{54.0\pm1.3}$ & $41.5\pm1.9$ & $58.1\pm1.0$ & $73.1\pm0.6$ \\
        & MuHDi~\cite{saporta2022multi} & $90.6\pm1.5$ & $\underline{80.0\pm0.5}$ & $81.6\pm0.0$ & $84.1\pm0.5$ & $\underline{86.9\pm1.3}$ & $65.4\pm2.8$ & $\underline{70.7\pm4.5}$ & $74.3\pm2.6$ & $\underline{80.6\pm4.0}$ & $51.0\pm8.0$ & $\bf{60.9\pm9.0}$ & $\underline{64.2\pm6.8}$ & $\underline{74.2\pm3.1}$ \\
        & GarDA (ours) & $92.6\pm0.1$ & $79.3\pm0.2$ & $\bf{86.3\pm0.1}$ & $\bf{86.1\pm0.1}$ & $86.3\pm0.4$ & $\bf{69.5\pm0.3}$ & $\bf{77.3\pm0.1}$ & $\bf{77.7\pm0.3}$ & $\bf{80.8\pm0.2}$ & $\bf{59.4\pm0.1}$ & $\underline{60.5\pm0.1}$ & $\bf{66.9\pm0.1}$ & $\bf{76.9\pm0.1}$ \\
        \midrule
        $\mathcal{D}_1$ & SD-UB & $-$ & $-$ & $-$ & $-$ & $91.0$ & $79.2$ & $85.0$ & $85.1$ & $-$ & $-$ & $-$ & $-$ & $-$\\
        $\mathcal{D}_2$ & SD-UB & $-$ & $-$ & $-$ & $-$ & $-$ & $-$ & $-$ & $-$ & $94.2$ & $82.2$ & $88.5$ & $88.3$ & $-$\\
        $\mathcal{D}_0 + \mathcal{D}_1 + \mathcal{D}_2$ & MD-UB & $94.1$ & $83.3$ & $89.7$ & $89.0$ & $92.3$ & $79.9$ & $88.4$ & $86.9$ & $92.9$ & $82.0$ & $88.2$ & $87.7$ & $87.9$ \\
        \bottomrule
    \end{tabular}
\end{sidewaystable}

\end{document}